\documentclass[conference]{IEEEtran}
\IEEEoverridecommandlockouts
\usepackage{cite}
\usepackage{amsmath,amssymb,amsfonts}
\usepackage{algorithmic}
\usepackage{graphicx}
\usepackage{textcomp}
\usepackage{xcolor}
\usepackage{cite}
\usepackage{amsmath,amssymb,amsfonts}
\usepackage{graphicx}
\usepackage{textcomp}
\usepackage{multirow}
\usepackage{xcolor}
\usepackage{subcaption}
\usepackage{float}
\usepackage{amsmath}
\allowdisplaybreaks
\usepackage{hyperref}
\hypersetup{
    colorlinks=true, 
    linkcolor=blue, 
    filecolor=magenta, 
    urlcolor=cyan, 
    citecolor=green 
}
\usepackage[ruled,linesnumbered]{algorithm2e}
\usepackage{graphicx}
\usepackage{subcaption}
\usepackage{float}
\usepackage{multirow}
\usepackage{amsmath,amssymb,amsfonts}
\usepackage[ruled,linesnumbered]{algorithm2e}
\usepackage{algorithmic}
\usepackage{textcomp}
\usepackage{xcolor}
\usepackage{marvosym}
\usepackage{hyperref}

\newcommand{\customlabel}[1]{\hypertarget{#1}{}}
\newcommand{\customref}[1]{\hyperlink{#1}{#1}}

\def\BibTeX{{\rm B\kern-.05em{\sc i\kern-.025em b}\kern-.08em
    T\kern-.1667em\lower.7ex\hbox{E}\kern-.125emX}}
\begin{document}

\title{Forgetting to Witness: Efficient Federated Unlearning and Its Visible Evaluation
}
\author{
\IEEEauthorblockN{Houzhe Wang\textsuperscript{1,2}, Xiaojie Zhu\textsuperscript{3(\Letter)}, Chi Chen\textsuperscript{1,2}}
\IEEEauthorblockA{
\textit{School of Cyber Security, University of Chinese Academy of Sciences, Beijing, China} \\
\textit{Institute of Information Engineering, Chinese Academy of Sciences, Beijing, China} \\
\textit{King Abdullah University of Science and Technology, Thuwal, Kingdom of Saudi Arabia  } \\
\{wanghouzhe, chenchi\}@iie.ac.cn \\
xiaojie.zhu@kaust.edu.sa
}
\thanks{(\Letter): Corresponding author}
}

\maketitle

\begin{abstract}
With the increasing importance of data privacy and security, federated unlearning has emerged as a novel research field dedicated to ensuring that federated learning models no longer retain or leak relevant information once specific data has been deleted.
In this paper, to the best of our knowledge, we propose the first complete pipeline for federated unlearning, which includes a federated unlearning approach and an evaluation framework.
Our proposed federated unlearning approach ensures high efficiency and model accuracy without the need to store historical data.
It effectively leverages the knowledge distillation model alongside various optimization mechanisms.
Moreover, we propose a framework named Skyeye to visualize the forgetting capacity of federated unlearning models. 
It utilizes the federated unlearning model as the classifier integrated into a Generative Adversarial Network (GAN).
Afterward, both the classifier and discriminator guide the generator in generating samples. 
Throughout this process, the generator learns from the classifier's knowledge. 
The generator then visualizes this knowledge through sample generation. 
Finally, the model's forgetting capability is evaluated based on the relevance between the deleted data and the generated samples.
Comprehensive experiments are conducted to illustrate the effectiveness of the proposed federated unlearning approach and the corresponding evaluation framework.
\end{abstract}

\section{Introduction}

Federated learning is a decentralized approach to machine learning that allows various clients to collaboratively train a global model without sharing their data \cite{konevcny2016federated} \cite{mcmahan2017communication}. 
This methodology effectively addresses a major challenge in conventional machine learning by enabling model training without relying on centralized data repositories and computational resources.

Nevertheless, in federated learning, clients may seek to remove their data contributions from the trained global model.
Additionally, recent regulatory frameworks like the European Union's General Data Protection Regulation (GDPR) \cite{regulation2018general} and the California Consumer Privacy Act (CCPA) \cite{pardau2018california} provide individuals with the authority to request the erasure of their personal data from any segment of the system within a reasonable timeframe.
It is crucial to note that even without direct data sharing, the global model could potentially infer information about the clients \cite{nasr2018comprehensive,song2020analyzing}. The predictions made by the global model might inadvertently reveal client-specific information \cite{salem2018ml}\cite{bagdasaryan2020backdoor}.
Therefore, there is an urgent need for a strategy to erase a client's data imprint from the trained global model.

The straightforward approach to ensure the global model forgets a particular client's input is to restart model training after excluding the user's data. However, this approach can be inefficient \cite{zhang2023fedrecovery}.
For example, the retraining strategy can become impractical in scenarios involving large datasets and complex models due to significant time, energy, computational costs, and scalability issues.

There is a growing interest in developing cost-effective machine unlearning algorithms to mitigate the effects of data deletion from trained models.
The approach is to remove the user's data from the model's parameters while maintaining the model's functionality \cite{bourtoule2021machine}. 
For example, 
Liu \textit{et al.} \cite{liu2021federaser} proposed a method for reconstructing a model that has undergone data loss by leveraging the parameter updates preserved on the server.
Izzo \textit{et al.} \cite{izzo2021approximate}, Neel \textit{et al.} \cite{neel2021descent}, and Wu \textit{et al.} \cite{wu2020deltagrad} have explored techniques that enable the server to approximate gradients efficiently during the unlearning process by leveraging historical gradients and model weights.
Due to the additional requirement of storing historical gradients, there is an extra cost for clients to maintain all historical data.


Moreover, quantifying the effectiveness of an unlearning algorithm in removing specific data remains a challenge in the above scenario. 
Ginart \textit{et al.} \cite{ginart2019making} proposed a metric similar to ($\varepsilon$, $\delta$)-differential privacy (DP). This metric verifies the indistinguishability between the outputs of the unlearning algorithm and those of a newly retrained model without the removed records.
Expanding on this concept, various certifications of unlearning have been proposed to validate the effectiveness of different mechanisms for unlearning data \cite{izzo2021approximate,guo2019certified,sekhari2021remember,neel2021descent,ullah2021machine}.
Unfortunately, researchers have uncovered the vulnerabilities in the above machine unlearning algorithms. Chourasia \textit{et al.} \cite{truedata} acknowledge the interdependence between the training data and the model, highlighting that the data utilized in training instills specific patterns within the model. Echoing Chourasia \textit{et al.}, Gupta \textit{et al.} \cite{gupta2021adaptive} argue that there are inherent flaws in the deletion certification approach of prior unlearning endeavors, which hinge on the indistinguishability between unlearned and retrained models.
Recent studies suggest that defining machine unlearning should focus on the algorithmic level, rather than solely on inspecting the model's parameters \cite{thudi2022necessity}. Meanwhile, research indicates that the model's parameters may remain unchanged even when specific data points are included or excluded during training \cite{shumailov2021manipulating}.
Another approach involves employing backdoors to verify compliance with data deletion requests by the model owner \cite{gao2024verifi}, \cite{sommer2022athena}. 
Specifically, these techniques assess the success rate of backdoor attacks on the model before and after unlearning with respect to backdoor samples. However, their effectiveness is probabilistic and constrained by the selection of data points designated as backdoors.
To verify the effectiveness of unlearning, users must locally establish backdoor data and actively monitor the success rate of backdoor attacks post-unlearning \cite{guo2023verifying}.
However, this verification method can only confirm that the model has unlearned the backdoor pattern. Still, it cannot prove that the model has completely removed the features associated with the samples containing the backdoor.
Moreover, the existing unlearning methods only focus on the models' final output, overlooking the internal feature maps. 
The impact of forgetting specific data on the model involves both adjusting the final output and modifying the intermediate layers. To minimize the effect on model accuracy and robustness, the modifications to the intermediate layers should be kept minimal.


To address the above challenges, in this paper, we first propose a federated unlearning approach that does not require historical data while offering better efficiency and considers both model output and internal attention map.  
Additionally, we introduce a federated unlearning evaluation framework named Skyeye. 
To the best of our knowledge, this is the first visible evaluation framework in the federated unlearning area. 


In the proposed federated unlearning approach, we employ a knowledge distillation model to facilitate the unlearning process. Specifically, the teacher model is an incompetent model, whereas the student model is the one required to unlearn the deleted data.
During the training process, the deleted data is input into the teacher model, and its output is used to guide the student model. By the end of the training, the student model has effectively learned the teacher model's inability to process the deleted data, thus achieving unlearning.
From this process, we can infer that the training process is efficient since it only involves the deleted data. Additionally, no extra historical data storage is required. To prevent disruptive changes to the model, attention map alignment is introduced to control the intermediate layers. Furthermore, learning rate decay, bounded loss, catastrophic forgetting prevention, and performance recovery mechanisms are incorporated to ensure model accuracy.


Inspired by Generative Adversarial Networks (GANs) \cite{goodfellow2014generative, yue2023gradient}, we propose the Skyeye framework to evaluate the effectiveness of federated unlearning algorithms through a visible approach.
Hitaj \textit{et al.} proposed leveraging GANs \cite{goodfellow2014generative} for inferring client data \cite{hitaj2017deep}. Yue \textit{et al.} demonstrated the use of GAN models to reconstruct images or data attributes of specific clients on the server \cite{yue2023gradient}.
All these examples illustrate that GANs have the capability to generate samples that closely resemble the distribution of training data. Essentially, GAN-generated models aim to capture and reproduce the statistical patterns and features present in the training dataset.
Motivated by this, our objective is to train a generator capable of capturing the knowledge acquired by a model. By analyzing the content generated by the generator, we aim to evaluate whether the model has completed the unlearning task.
To achieve this goal, we propose the Skyeye framework, which utilizes GAN technology.
The Skyeye framework consists of three main phases:
1) Classifier Integration: This phase involves integrating the classifier, which has undergone the unlearning process, into the GAN-based framework. The classifier serves as the third component alongside the discriminator and generator.
2) Training: In this phase, the framework extracts knowledge from the classifier and guides the generator, alongside the discriminator, to learn from this extracted knowledge. The training process ensures that the generator captures and reproduces the learned information effectively.
3) Decision-making: The final phase assesses the effectiveness of the unlearning process. It evaluates whether the model has effectively forgotten the deleted data by examining the content generated by the generator. If the generator can generate content closely resembling the deleted data, the unlearning algorithm is deemed ineffective; otherwise, it is considered successful.
This structured approach enables Skyeye to effectively evaluate the efficacy of federated unlearning algorithms and visualize the unlearning process.





Overall, the main contributions of this paper can be summarized as follows.
\begin{enumerate}
    \item We propose an efficient federated unlearning approach that ensures both model accuracy and robustness, without the need for additional historical data storage. 
    By utilizing a knowledge distillation model, minimizing adjustments to internal attention maps, and incorporating various optimization modules, the experiments show that our method processes deleted data with minimal impact on the accuracy of the remaining data.
    \item We introduce a new paradigm to evaluate the effectiveness of federated unlearning algorithms by leveraging the data recovery capabilities of GANs. 
    Our proposal integrates the unlearned model into a GAN framework to attempt the recovery of deleted data, providing a visible method to assess the effectiveness of the unlearning capability based on the recovered results.
    \item We introduce Skyeye, to the best of our knowledge, the first federated unlearning evaluation framework designed to assess the effectiveness of various machine unlearning algorithms. Additionally, comprehensive experiments have been conducted, demonstrating its efficacy. 
    Our code is available at \href{https://anonymous.4open.science/r/Forgetting-to-Witness-B603/}{https://anonymous.4open.science/r/Forgetting-to-Witness-B603.}
    
\end{enumerate}
\section{Related Work}
In federated unlearning tasks, the objective is to eliminate the specific knowledge learned from users' local data within the trained global model.
The existing federated unlearning methods can be broadly classified into two types: exact unlearning and approximate unlearning.
\subsection{Exact Unlearning}
Exact unlearning is based on the method of retraining from scratch. It mainly improves the training process of the model, so that when the model needs to forget data, it can reduce the computational and time costs of model training.

Bourtoule \textit{et al.} \cite{bourtoule2021machine} introduced SISA training as an approach to alleviate the computational costs associated with forgetting.
Building upon the SISA framework, several related studies have been proposed.
The random forest algorithm improves the performance of the model by building multiple decision trees and summarizing their predictions \cite{brophy2021machine}. 
In terms of unlearning, each tree corresponds to a slice in the SISA framework, 
trained independently and isolated from the influence of data points, 
so that when it is necessary to forget specific data points, only those trees that include the data point need to be retrained.
DC-k-means \cite{ginart2019making}, as an extension of k-means, uses a tree-based hierarchical clustering method to achieve exact unlearning. It randomly divides the data into multiple subsets and trains a k-means model on each subset, finally constructing the final clustering result by merging these models.
KNOT \cite{su2023asynchronous} uses the SISA framework to implement client-level asynchronous exact unlearning. Through cluster aggregation, clients are grouped, and the server aggregates the model within the cluster, while each cluster trains independently. Data deletion requests only trigger retraining of clients in the same cluster.

Additionally, there are researchers exploring the relationship between the model and data, as well as efforts focused on enhancing training efficiency.
\cite{cao2015towards} draw inspiration from statistical query learning and design an intermediate layer called "Summation", which serves as a buffer between the machine learning algorithm and the training data, making the algorithm learn through summarized statistical information rather than original data.
Liu et al. 
\cite{liu2022right} use the first-order Taylor expansion approximation technique to customize a rapid retraining algorithm based on diagonal experience FIM.
\subsection{Approximate Unlearning}
Approximate unlearning aims to minimize the impact of data that needs to be deleted or forgotten to an acceptable level, while also achieving an efficient unlearning process.

Compared to exact unlearning techniques, approximate unlearning offers several advantages, including better computational efficiency, lower storage costs, and greater flexibility.
In terms of computational efficiency, approximate unlearning methods reduce computational costs by minimizing rather than completely deleting the impact of data, as opposed to exact unlearning methods that require retraining with the remaining data \cite{guo2019certified}. For instance, the method proposed in \cite{guo2019certified} adjusts model parameters to reduce the influence of specific data, thus reducing computational intensity compared to exact unlearning.
Regarding storage overhead, approximate unlearning methods, such as those presented by Sekhari \textit{et al.} in \cite{sekhari2021remember}, store only necessary statistical information of the data, thereby significantly reducing storage costs.
In terms of flexibility, approximate unlearning methods are highly adaptable, as they typically do not rely on specific learning models or data structures, allowing for broader application to a variety of learning algorithms \cite{cao2015towards}. This flexibility is related to the trade-off between completeness and efficiency made by approximate unlearning, allowing for adaptation to new data and tasks by accelerating the unlearning process and reducing costs while maintaining model performance. For example, 
Zhang \textit{et al.} \cite{zhang2023fedrecovery} eliminate client influence by extracting the weighted sum of gradient residuals from the global model and introducing Gaussian noise. This process is designed to achieve statistical indistinguishability between the unlearned and retrained models.
Liu \textit{et al.} \cite{liu2021federaser} reconstruct the forgotten model using parameter updates stored on the server, introducing a novel calibration method to adjust client updates. This innovative approach aims to enhance the speed of unlearning while preserving model performance.
Additionally, Baumhauer \textit{et al.} \cite{baumhauer2022machine} and Thudi \textit{et al.} \cite{thudi2022necessity} emphasize the pursuit of higher efficiency in machine unlearning by relaxing the requirements for both effectiveness and provability.
Izzo \textit{et al.} \cite{izzo2021approximate}, Neel \textit{et al.} \cite{neel2021descent}, and Wu \textit{et al.} \cite{wu2020deltagrad} explore techniques for the server to effectively approximate gradients during the unlearning process by leveraging historical gradients and model weights.
Chourasia \textit{et al.} \cite{chourasia2023forget} enhances model robustness in addressing data deletion. Wang \textit{et al.} \cite{wang2024goldfish} propose Goldfish, an efficient federated unlearning framework designed with four modules: the basic model, loss function, optimization, and extension. Each module is  crafted to enhance the framework's practicality and ensure seamless integration into real-world federated learning scenarios.

\subsection{ Evaluation Methods of Federated Unlearning }

Despite the various methods proposed for evaluating the effectiveness of federated unlearning, there remains a lack of approaches capable of visualizing the outcomes of unlearning.

Ginart \textit{et al.} \cite{ginart2019making} introduced a novel metric inspired by 
 \(\varepsilon\), \(\delta\)-differential privacy (DP). This metric assesses the degree of indistinguishability between the outputs of an algorithm intended for ``unlearning" information and those of a model that has been freshly retrained while excluding the data involved in the unlearning process.
Expanding upon this foundational idea, several methodologies have been proposed to certify data erasure and substantiate the efficacy of various unlearning mechanisms \cite{izzo2021approximate, guo2019certified, sekhari2021remember, neel2021descent, ullah2021machine}. However, researchers have identified vulnerabilities in the mentioned machine unlearning algorithms. Chourasia \textit{et al.} \cite{truedata} emphasizes the interdependence between training data and models, highlighting that the patterns embedded within models originate from the characteristics of the training data. Echoing Chourasia \textit{et al.}, Gupta \textit{et al.} \cite{gupta2021adaptive} argue that there are inherent flaws in the deletion certification approach of prior unlearning endeavors, which hinge on the indistinguishability between unlearned and retrained models.

Another approach being explored involves using backdoors to enforce compliance with data deletion requests as mandated by the model owner.
These techniques specifically evaluate the success rate of backdoor attacks on the model's performance before and after the unlearning process, focusing on the samples identified as backdoors. However, it is important to note that the effectiveness of these techniques is probabilistic and depends on the selection of data points designated as backdoors.
To effectively verify the unlearning process, users must establish local backdoor datasets and actively monitor the success rate of backdoor attacks post-unlearning \cite{guo2023verifying}. 
\subsection{GANs Employed in Data Reconstruction }
GANs consist of two primary components: a generator and a discriminator, as initially outlined by Goodfellow \textit{et al.} \cite{goodfellow2014generative}.
The generator's goal is to create synthetic data that closely matches the distribution of the training dataset, while the discriminator's role is to distinguish between genuine and generated samples.
This adversarial process continues until the discriminator can no longer reliably distinguish between real and fake samples, indicating that the generator has achieved a high level of realism in its output \cite{goodfellow2014generative, qu2019gan, mirza2014conditional}.
In the context of federated learning, an attacker can exploit GANs to mimic data contributions from other participants, potentially enabling unauthorized access to sensitive information \cite{wang2019beyond, zhang2019poisoning, fang2021privacy}. 

\section{Preliminaries}
In this section, we provide background knowledge on machine learning and federated learning. For additional preliminary information, please refer to Appendix \customref{A}.

\subsection{Machine Unlearning}

Machine unlearning refers to the process of eliminating the influence of specific training data points on a pre-trained machine learning model \cite{chen2021machine}. Specifically, given a model with parameters $w^\ast$ trained using a learning algorithm $A$ on a dataset $D$, and a subset $D_f\ \subseteq\ D$ to be removed, the machine unlearning algorithm $U(A(D),\ D,\ D_f)$ aims to obtain a new model with parameters $w^-$ that eliminates the influence of $D_f$ while maintaining performance on the remaining dataset $D_r = D \backslash D_f$.

\subsection{Federated Learning}
Federated Learning (FL) \cite{mcmahan2017communication} is an innovative distributed machine learning paradigm that has recently garnered significant attention. It allows individuals to collaborate in training a global machine learning model without the need to share their private training data with others. FL typically consists of a server and \(N_{users} \) users. During the FL training process, at the initial stage, each user \( i \) initializes its local user model \( \omega_i^0 \) using the initialized global model \( \omega_0 \). In the subsequent rounds, each user \( i \) conducts local training using the current round \( t \) global model \( \omega_t \) and its local training data \( D_i \), then sends its local model update \( \omega_i^{t+1} \) to the server. After receiving model updates from all users, the server utilizes a specific aggregation rule to combine the received model updates and further update the global model. The updated global model \( \omega_{t+1} \) is then distributed to all users for the next round of training.

For instance, the FedAvg \cite{mcmahan2017communication} aggregation rule calculates the average of model updates to obtain the global model, which is used in non-adversarial scenarios. One advantage of FL over centralized learning is that users no longer need to send their private training data to the server. The model aggregation scheme of FedAvg is shown as equation \ref{Fed联邦聚合公式}:
\begin{equation}
    \omega_{t+1}=\frac{1}{N_{users}}\sum_{i\in N_{users}}\omega_i^{t+1}
    \label{Fed联邦聚合公式}
\end{equation}

\section{Proposed Visible Federated Unlearning Approach}

In this section, we first present the proposed federated unlearning approach, detailed in Algorithm \ref{alg:Unlearned} of Appendix \customref{B}. Afterward, we demonstrate the visible evaluation algorithm for assessing data forgetting ability, detailed in  Algorithm \ref{alg:training-phase} and \ref{Decision-making phase} of Appendix \customref{B}. Finally, we discuss potential improvements.

\subsection{Proposed Federated Unlearning Approach}



The proposed federated unlearning approach primarily comprises two components: basic model selection and loss function construction. The basic model determines the unlearning approach and the loss function influences the quality of the training process.
In the basic model, while we employ a knowledge distillation model, we adopt an incompetent teacher model \cite{chundawat2023can}. 
The incompetent teacher model eliminates the need for datasets by guiding the student model to forget specific data locally, thereby reducing the risk of data leakage and enhancing the efficiency of data deletion. 
The goal of the unlearning algorithm is to ensure that when the model is presented with the deleted data again, it behaves as if it has never encountered this data before.
To achieve this, we employ an incompetent teacher model to guide the student model in learning the removed data.
The purpose is to obtain a student model that performs randomly in predicting the deleted data. 
In particular, the knowledge distillation process between the teacher model and the student model occurs during the unlearning process.
Moreover, inspired by Gong \textit{et al.} \cite{gong2023redeem}, our loss function additionally considers the discrepancy of internal layers in addition to the output layers. 
The objective is to align the attention maps of the feature maps output by the student and teacher models in the intermediate layers of the network. 
When a model is subjected to a backdoor attack, there is a discernible change in the attention maps output by the intermediate layers during the forward pass. These changes correlate the model's final output with the pixel values in the region where the backdoor trigger is located.
Therefore, we consider the intermediate results of the model rather than focusing solely on the final output.

\begin{figure}
    \centering
    \includegraphics[width=1\linewidth]{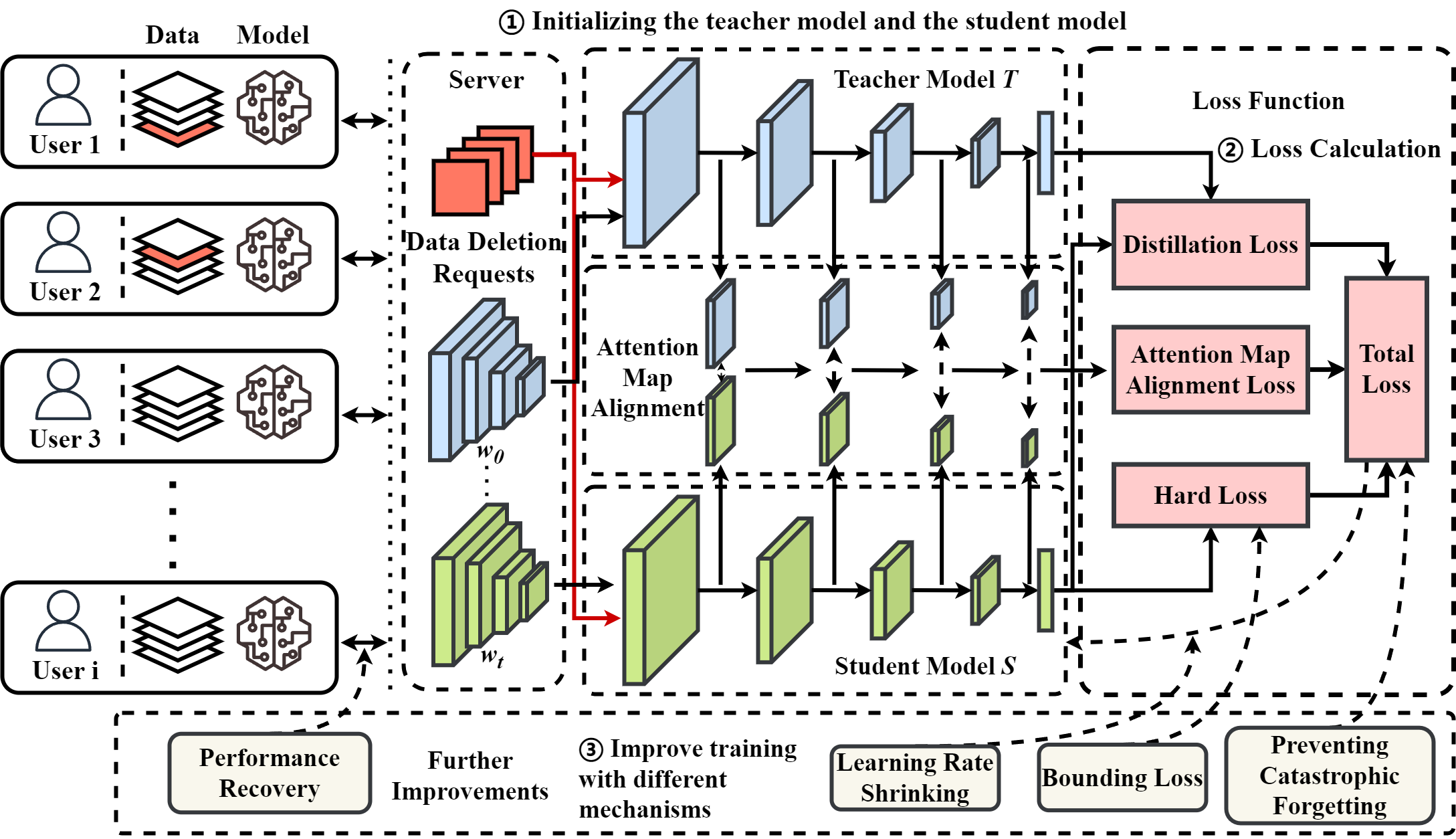}
    \caption{Proposed federated unlearning approach. 
      When the server receives a user's deletion request, the following steps are executed sequentially on the server side:
     \textcircled{1} Initialization of the teacher model and the student model. It takes the current global model as the student model and an incompetent model as the teacher model.
     \textcircled{2} Loss Calculation. 
     The total loss consists of distillation loss, attention map alignment loss, and hard loss.  
     \textcircled{3} Improvements of training with different mechanisms. To further enhance the training process, we propose mechanisms for handling loss explosion, preventing catastrophic forgetting, and ensuring performance recovery. 
}
    \label{去学习的流程图}
\end{figure}

To elaborate on the proposed unlearning approach, we first present an example and then provide a detailed explanation.
For instance, during the 
$t$-th round of training,  a user $i$ issues a data deletion request. 
Based on this request, the local dataset 
$D_i$ of the user can be divided into a forgetting dataset $D_f^i$ and a remaining dataset $D_r^i=D_i\setminus D_f^i$. 
The server collects all users' local deletion data $D_f^i$, $i \in [1,N_{users}]$, and obtains the users' deletion requests $D_f$, then proceeds with the unlearning process.
Specifically, the student model is initialized with the global model, and the teacher model is configured with an incompetent model. The forgetting dataset $D_f$ is input into the teacher model, and its output is used to guide the student model.
To precisely define the concept of guidance, we formulate the loss function as shown in equation \ref{loss_function},
where the unlearning loss \(\mathcal{L}_{unlearn}\) is defined as the weighted sum of three components: the distillation loss \(\mathcal{L}_D\) between the teacher and student model outputs, the attention map alignment loss \(\mathcal{L}_{AMA}\) between the teacher and student models, and the hard loss \(\mathcal{L}_H\) between the output of the student model and the ground truth on the removed data. The constants \(\mu_D\), \(\mu_{AMA}\), and \(\mu_H\) are parameters used to balance the weights of the different loss components. The definition of those loss components are further explained below. 
\begin{equation}
    \mathcal{L}_{unlearn} = \mu_D \mathcal{L}_D + \mu_{AMA} \mathcal{L}_{AMA} + \mu_H \mathcal{L}_H
    \label{loss_function}
\end{equation}

\textit{Distillation Loss}. 
The \( \mathcal{L}_D \) loss measures the discrepancy between the outputs of the teacher and student models. We utilize the teacher model's outputs as the labels for the student model. The output vector from the teacher model is transformed into a predicted confidence vector via the softmax function. The confidence of the teacher model \( T \) for a sample \( x \) is represented as \( P_{x}^T \), which is calculated through the following equation \ref{loss_dis_teacher}:
\begin{equation}
     P_{x}^T = \frac{\exp\left(\frac{v_i}{Temp}\right)}{\sum_{j=1}^{C}{\exp\left(\frac{v_j}{Temp}\right)}} 
     \label{loss_dis_teacher}
\end{equation}
where \( Temp \) represents the distillation temperature, \( v_i \) denotes the teacher model \( T \)'s prediction for \( x \) being the correct label \( i \) (with \( i \in \left[1, C \right] \), $C$ denotes the total number of labels), and \( v_j \) signifies the teacher model \( T \)'s prediction for \( x\) being labeled as \( j \) (with \( j \in \left[1, C \right] \)). Using a similar method, we define the confidence \( P_{x}^S \) of the student model \( S \) for the sample \( x \), which can be calculated by the following equation \ref{loss_dis_student}:
\begin{equation}
    P_{x}^S = \frac{\exp\left(\frac{z_i}{Temp}\right)}{\sum_{j=1}^{C}{\exp\left(\frac{z_j}{Temp}\right)}}
    \label{loss_dis_student}
\end{equation}
where \( z_i \) represents the confidence with which the student model \( S \) predicts that \( x \) is labeled with the correct label \( i \) (where \( i \in \left[1, C \right] \)), and \( z_j \) indicates the confidence with which the student model \( S \) predicts \( x \) labeled \( j \) (where \( j \in \left[1, C \right] \)). Ultimately, we define the distillation loss \( \mathcal{L}_D \) as follows:
\begin{equation}
    \mathcal{L}_D = -\sum_{x \in D_f} P_{x}^T \log P_{x}^S
\end{equation}
The formula illustrates that the larger the discrepancy between the prediction distributions of the teacher and student models on the forgotten dataset, the greater the loss value.

\textit{Attention Map Alignment Loss.} 
Existing machine unlearning methods \cite{xu2024machine,chundawat2023can,wu2022federated} focus solely on aligning the model's predictive results, while neglecting how the model processes data internally.
It significantly weakens the model's robustness. 
The attention map alignment mechanism we propose is designed to mitigate this weakness. 
We focus not only on the model's final output but also on every step of the data processing within the model. 
Our attention alignment mechanism provides knowledge-free attention maps through an incomplete teacher model. 

The \( \mathcal{L}_{AMA} \) loss measures the discrepancy in attention maps between the teacher and student models under the same input. 
Given the teacher model \( T \), we define the output feature map of the teacher model at layer \( l \) as \( T^l \in \mathbb{R}^{C_l \times H_l \times W_l} \), where \( C_l \), \( H_l \), and \( W_l \) represent the number of channels, the height, and the width of the attention map, respectively. 
We define an attention map computation function \( A: \mathbb{R}^{C_l \times H_l \times W_l} \rightarrow \mathbb{R}^{H_l \times W_l} \) that reduces the channel dimension of the feature map \( T^l \) as equation \ref{attention_map_compute}:
\begin{equation}
    A_{\text{sum}}^q(S^l) = \sum_{i=1}^{C_l} \left| T_i^l \right|^q
    \label{attention_map_compute}
\end{equation}
where \( T_i^l \in \mathbb{R}^{H_l \times W_l} \) is the feature slice of the \( i \)-th channel in the \( l \)-th layer feature map of the teacher model \( T \),  $\left| T_i^l \right|$ is its absolute value and \( q~(q \geq 1) \) is a constant. 
In the attention map computation function, we use the absolute value of each channel feature slice for a weighted sum, where a larger \( q \) value gives more weight to the sum of regions with larger absolute values. We calculate the attention map of the student model \( S \) in the same way. We denote the difference in attention maps generated by the student model and the teacher model for the sample \( x \) as equation \ref{注意力图层层计算}:
\begin{equation}
    dis\left(T^l,S^l;x\right) = \left \| \frac{A\left(T^l\left(x\right)\right)}{\left \| A\left(T^l\left(x\right)\right) \right \|_2 } - \frac{A\left(S^l\left(x\right)\right)}{\left \| A\left(S^l\left(x\right)\right) \right \|_2 }\right \|_2
    \label{注意力图层层计算}
\end{equation}
 where $\left \| \cdot \right \|_2$ is the $L_2$ norm.
We calculate the difference between normalized attention maps and ultimately define \( \mathcal{L}_{AMA} \) as:
\begin{equation}
    \mathcal{L}_{AMA} = \frac{1}{\left| D_f \right|} \sum_{x \in D_f} dis(T^l, S^l; x)
    \label{注意力图损失}
\end{equation}

The equation \ref{注意力图损失} computes the average discrepancy of the attention maps between the teacher and student models over the forgetting dataset \( D_f \), where \( \left| D_f \right| \) denotes the number of elements in \( D_f \), and \( dis(T^l, S^l; x) \) represents the difference measure for a given sample \( x \).

\textit{Hard Loss. }
Hard Loss represents the negative loss value that a model incurs on data that has been removed or forgotten. The idea is that during normal training, the model learns the knowledge of samples by minimizing the loss across all samples. When the model is required to unlearn a particular sample it has previously learned, it should maximize the loss it incurs on that sample. Therefore, we define the loss of the sample using a cross-entropy loss function, as shown 
in equation \ref{eq:hard_loss}, where, \(S(x)\) represents the confidence with which the student model predicts the input feature vector \(x\) as its corresponding \(y\). Unlike the training process, \(\mathcal{L}_H\) aims to reduce the accuracy of the student model on the forgotten samples.
\begin{equation}
\label{eq:hard_loss}
    \mathcal{L}_H = \sum_{(x, y) \in D_f} y \log S(x)
\end{equation}
\subsection{Evaluation Framework of Federated Unlearning}

\begin{figure*}
    \centering
    \includegraphics[width=1\linewidth]{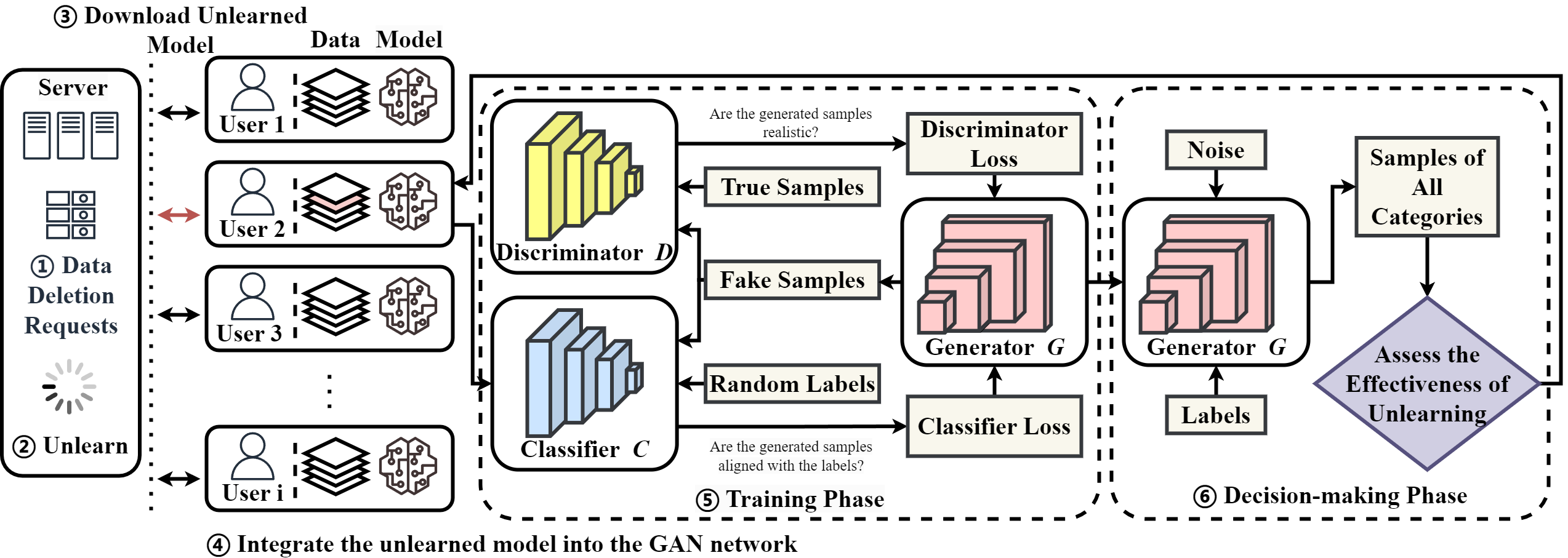}
    \caption{Skyeye framework of forgetting capability evaluation. The main procedures are described below. 
     \textcircled{1} Users send data deletion requests to the server. 
    \textcircled{2} Server executes the unlearning algorithm. 
    \textcircled{3} Users download the unlearned model. 
    \textcircled{4} Integrate the unlearned model into the GAN network.
    \textcircled{5} Training Phase. The generator takes random noise samples and random labels as input to generate fake samples, while the discriminator is trained to differentiate between real and fake samples.
    The classifier receives the same random labels and fake samples to classify them into different categories. Ultimately, both the classifier and discriminator work together to guide the generator in generating samples. 
    \textcircled{6} Decision-making Phase. Users evaluate the server's unlearning algorithm by assessing the relevance between the images generated by the generator and the deleted data.
    }
    \label{验证方法流程图}
\end{figure*}

As shown in Fig.\ref{验证方法流程图}, our evaluation framework consists of three components: generator, discriminator, and classifier. 
Initially, when a user requests the deletion of their data, the server deploys the unlearning algorithm.  
Subsequently, when the user receives the unlearned model, the user decides  whether to verify the effectiveness of the received federated unlearning model. If the user decides to verify the unlearning model, the training process of the evaluation framework begins to be executed locally on the user side.
The complete training phase is detailed in  Algorithm \ref{alg:training-phase} of Appendix \customref{B}. Finally, the evaluation process is executed during the decision-making phase, as outlined in Algorithm \ref{Decision-making phase} of Appendix \customref{B}. 


Specifically, during the training process, the generator pursues two optimization objectives. Firstly, it aims to maximize the confidence level of pseudo-samples being classified as real by the discriminator, thereby enhancing the realism of the generated images.
Secondly, it aims to maximize the confidence level of pseudo-samples being correctly classified into their respective categories, thus effectively learning category-specific information from the classifier.
Ultimately, the ideal outcome is that the pseudo-samples generated by the generator should deceive the discriminator while still adhering to the classifier’s criteria.



Let us consider the MNIST dataset as an example to demonstrate how the evaluation framework operates: the original classifier is trained on the complete MNIST dataset, and after applying the specified unlearning mechanism, the post-unlearning classifier removes information related to the category of the digit `3’. Subsequently, this post-unlearning classifier is integrated into the GAN network to train both the generator and discriminator. Ultimately, the resulting generator can generate high-quality images of the other nine digit categories, excluding `3'.

In particular, we first train the original classification model \( C_o \) on the entire users' dataset \( D \). 
\( C_o \) outputs a predicted label \(\hat{y} \in \left[1, c\right] \) for each sample \( x \in D \), where \( c \) is the number of classes contained in the dataset \( D \). 
Due to the users' request to forget the data \( D_f \), \( D \) is partitioned into two disjoint subsets: the forgetting dataset \( D_f \) and the remaining dataset \( D_r \), where \( D_r = D \setminus D_f \). 
After executing the unlearning mechanism \( M \), we obtain the unlearned classification model \( C_u = M(C_o, D_f) \). 
Ideally, \( C_u \) should not contain any information related to the dataset \( D_f \). To visualize the knowledge retained by \( C_u \), \( C_u \) is incorporated as a classifier into the training process of a GAN network:

\begin{itemize}
    \item The generator \( G \) accepts a random vector \( z \in \mathbb{R}^d \) and a random label \( y_r \in \left[1, C\right] \) as inputs and correspondingly outputs a pseudo-samples \( \widetilde{x} = G(z, y_r) \).
    \item The discriminator \( D \) accepts either a real sample \( x \) or a pseudo-sample \( \widetilde{x} \) as input and outputs the confidence level \( p = D(x) \) that it classifies the input as a real sample.
    \item The classifier \( C_u \) accepts pseudo-samples as input and outputs a class prediction vector \( C_u
    (\widetilde{x}) \) for the pseudo-samples.
\end{itemize}

During the training process, the classifier solely provides output results for each input sample. However, the generator \( G \) and the discriminator \( D \) are trained alternately based on their respective loss functions. The loss function for the generator is defined as equation \ref{生成器的目标函数}:
\begin{equation}
   l_G = -\mu_d \log \left( D \left( G \left( z, y_r \right) \right) \right) - \mu_c y_r \log C \left( G \left( z, y_r \right) \right)
    \label{生成器的目标函数}
\end{equation}
where \( -\log \left( D \left( G \left( z, y_r \right) \right) \right) \) is the loss generated by the generator \( G \) with respect to the discriminator \( D \). The objective is to maximize the confidence level with which  the discriminator classifies the pseudo-samples generated by \( G \) 
 as real, thereby aligning the sample distribution generated by \( G \) more closely with that of the real samples.
\( - y_r \log C \left( G \left( z, y_r \right) \right) \) represents the loss generated by the classifier \( C_u \). The goal is to minimize the cross-entropy loss function, \textit{i.e.}, to maximize the confidence with which the classifier \( C_u \) classifies the pseudo-samples generated by
\( G \) as \( y_r \). This ensures that the generated pseudo-samples are more consistent with the classifier’s perception.
 \( \mu_d \) and \( \mu_c \) are parameters used to balance the importance of the two components of the loss function. { They are all set to 0.5.}

The objective of the discriminator is to differentiate between the pseudo-samples generated by the generator \( G \)  and the real samples.
The corresponding loss function is defined in equation \ref{判别器损失}. Minimizing \( l_D \) involves maximizing the discriminator's confidence in identifying real samples as real while minimizing its confidence in classifying the generator's pseudo-samples as real.
The parameters \( \mu_f \) and \( \mu_r \) are used to weight the importance of the two components of the loss function.
\begin{equation}
    l_D = -\mu_f \log \left( 1-D \left( G \left( z, y_r \right) \right) \right) - \mu_r \log \left(  D \left( x \right) \right)
    \label{判别器损失}
\end{equation}
\subsection{Further Improvements}
\subsubsection{Handling Loss Explosion}
When a model needs to forget a sample \( x \), it involves increasing the model's loss 
\( \mathcal{L}(x) \) on that sample. This intentional increase aims to reduce the model's accuracy in predicting \( x \). However, without constraints on the value of 
\( \mathcal{L}(x) \), excessively maximizing it might negatively impact the overall performance of the model.

To effectively forget the removed data, during the training process,  the model needs to prioritize increasing the loss on the deleted samples while minimizing the overall loss function.
This approach may inadvertently increase the loss on the remaining dataset, potentially leading to a decline in the model's performance. 
To address the above scenario, we employ two techniques: bounded loss and learning rate decay.

\textit{Bounded loss. }
The purpose of the bounded loss is to limit the increase of the loss \(\sum_{x \in D_f}\mathcal{L}(x)\) when minimizing the loss function, thus preventing it from growing without bound. Specifically, we use the ReLU (Rectified Linear Unit) function and a maximum loss bound constant \(BND\) to constrain the loss, modifying \(\mathcal{L}_{H}\)  of \(\mathcal{L}_{unlearn}\) to:
\begin{equation}
    \mathcal{L}_H = \sum_{x \in D_f} \text{ReLU}\left(BND - \mathcal{L}(x)\right)
\end{equation}
When \(\mathcal{L}(x) \geq BND\), the value of \(\text{ReLU}\left( (BND - \mathcal{L}(x))\right)\) is always zero, which means that the upper limit for each \(\mathcal{L}(x)\) is \(BND\). The selection of the constant \(BND\) is closely tied to the loss function used in the experiments. 
Moreover, the value of \(BND\)  signifies the intensity of the user's desire to forget. A larger 
\(BND\)  indicates a stronger emphasis on forgetting.

\textit{Learning rate decay.}
Learning rate decay is a technique employed to assist the model in gradually approaching a global convergence point during regular training.
By gradually reducing the learning rate, we use a smaller step size to search for the convergence point, which results in progressively smaller loss values and correspondingly smaller gradient updates. 
In unlearning tasks, attempting to maximize the loss of removed data at the same learning rate can accelerate the increase in loss values.
In such a scenario, where the bounded loss constraint mentioned above is satisfied, the actual \(\mathcal{L}(x)\) may still be significantly less than \(BND\). 
To ensure that \(\mathcal{L}(x)\) and \(BND\) converge closely by the end of the optimization process, it is essential to significantly reduce the learning rate during unlearning.


Bounded loss restricts how much the loss can increase for a specific sample, preventing the model from excessively prioritizing individual samples at the expense of overall performance. In contrast, learning rate decay progressively reduces the step size during optimization iterations, aiding in fine-tuning model parameters without drastically increasing loss on other samples. These methods collaborate to enable the model to unlearn specific data points while maintaining stability and overall performance.

\subsubsection{Preventing Catastrophic Forgetting}
During the unlearning process, the model seeks a new convergence point in its parameter space guided by the loss function. Without constraints on this search space, the model may find a new point far from the original, potentially causing it to forget previously learned knowledge.

A direct approach to mitigate catastrophic forgetting is to incorporate all past data into the training set for each model update. However, this method becomes impractical as the training set size increases over time, leading to longer model update times.

Instead of using all the past data, we choose to add a regularization term \( \| \omega - \omega_o \| ^2\) to the $\mathcal{L}_{unlearn}$. 
This term penalizes deviations between the current model parameters 
\(\omega\) and the original model parameters 
\(\omega_o\) during the unlearning process. 
By keeping the updated model close to the original, this regularization helps balance performance with the unlearning objective.

\subsubsection{Performance Recovery}
After the server completes the unlearning task, the unlearned global model will be distributed to users. 
The model will first be tested on the local dataset of each user.
It will then undergo local training on the dataset of the user with the lowest test performance, and the resulting model will be sent to the server. 
The server will then forward it to other users.
The entire procedure is repeated until the performance of the distributed model meets the minimum required standard.

\section{Experiment}

In this section, we evaluate the effectiveness of the proposed unlearning approach and its evaluation framework using different datasets and models. 
We begin by clarifying the objectives of the experiment. 
Next, we introduce the experimental setup, which includes an overview of the datasets and models used, a description of the baselines for comparison, and the hyperparameter settings. 
Finally, we present and analyze the experimental results.

\subsection{Experimental Goal}
For the experiments of the unlearning approach we proposed, we first verified the effectiveness of the proposed approach, that is, whether it can achieve the forgetting of data. 
Next, we need to explore the impact of the model's hyperparameter settings on the experimental results. 
We also need to investigate the influence of different loss function components on the final results to assess the effectiveness of each part of the loss. 
Additionally, we should divide the unlearning task into sample unlearning and category unlearning to compare the effectiveness of the model under the two types of unlearning tasks. 
At the same time, we should also select appropriate baselines for comparative experiments.

Regarding the evaluation framework we proposed, we need to verify whether the evaluation framework is truly effective, that is, whether the evaluation framework can correctly reflect the information contained in the model. 
And in the context of federated learning, can users with partial data evaluate the forgetting of local data? 
At the same time, we need to conduct ablation studies to evaluate the role and effectiveness of each part of the loss function during the training process.

\subsection{Experimental Setup}

In the experiment, all models are implemented using PyTorch and executed on two machines: one equipped with an NVIDIA 1660ti GPU and another with an NVIDIA 3060 GPU.

\textbf{Dataset Description}.
In the experiments, we utilized four publicly available ML datasets: MNIST \cite{lecun1998gradient} , AT\&T \cite{samaria1994parameterisation}, CIFAR-10 \cite{krizhevsky2009learning}, and CIFAR-100 \cite{krizhevsky2009learning}. As shown in Table \ref{tab:dataset}, these datasets encompass varying attributes, dimensions, and the number of categories. 
For example, the version of the AT\&T dataset we used is composed of 400 images of 64×64 pixels. This dataset contains images of 40 different individuals, that is, 10 images per person.
In the setting of the FL environment, we uniformly distribute the data from the training datasets to all users.

\begin{table}[]
\centering
\renewcommand\arraystretch{1}
\tabcolsep=0.2cm
\caption{Dataset Description}
\label{tab:dataset}
\begin{tabular}{ccccc}
\hline
Dataset       & Dimensions & Classes & Training & Test  \\ \hline
MNIST         & 784        & 10      & 60000    & 10000 \\
AT\&T & 4096        & 40      & 360    & 40 \\
CIFAR-10      & 3072       & 10      & 50000    & 10000 \\ 
{CIFAR-100}     & {3072}       & {100}     & {50000}    & {10000} \\ \hline
\end{tabular}
\end{table}

\textbf{Models}.
The MNIST dataset, with its images of handwritten digits, is well-suited for the traditional LeNet-5 model due to its simplicity and relatively low dimensionality. In contrast, for more complex image datasets like CIFAR-10 and CIFAR-100, deeper networks such as ResNet are better at capturing intricate features.
Following the most related works \cite{chundawat2023can,hitaj2017deep, gong2023redeem}, we adopt four different models for our evaluation. 
In particular, the model for MNIST is a traditional LeNet-5 model \cite{liu2022right} \cite{lecun1998gradient}, which consists of 2 convolution layers, 2 max pool layers, and 2 fully connected layers for prediction output.
The model for AT\&T is a modified LeNet-5 model \cite{liu2022right} \cite{lecun1998gradient}, which consists of 1 convolution layer, 1 max pool layer, and 2 fully connected layers for prediction output.
The model selected for CIFAR-10 is ResNet32, a variant of the Residual Network (ResNet) architecture proposed by He \textit{et al.} \cite{he2016deep}. It comprises 32 layers, structured with multiple residual blocks that enable the network to learn residual functions relative to its input. 
The model chosen for CIFAR-100 is ResNet56, an advanced variant of ResNet architecture introduced by He \textit{et al.} \cite{he2016deep}. This model, tailored for the CIFAR-100 dataset, comprises 56 layers with multiple residual blocks. 

Due to the inherent differences between the AT\&T dataset and the MNIST dataset, we adopt the methodology in the previous work by Hitaj et al. \cite{hitaj2017deep} for constructing the GAN networks.
For a detailed description of the model architectures, please refer to the Appendix \customref{C}.

\textbf{Hyperparameters}.
Our hyperparameter settings are aligned with those of Hitaj \textit{et al.} \cite{hitaj2017deep}.
For the MNIST dataset, we utilize a batch size of 100, a learning rate of 0.01, and the SGD optimizer.
For the CIFAR-10 and CIFAR-100 datasets, we use a batch size of 128, a learning rate of 0.01, and the Adam optimizer.
Regarding the deletion rate, consistent with \cite{wang2023machine}, we evaluate the values at 2\%, 4\%, 6\%, 8\%, 10\%, and 12\%.
In the category unlearning task, following \cite{chundawat2023can}, we set the number of categories to be forgotten to either 1 or 20\% of the total category count.
In addition, following the recommendations from previous experiments \cite{li2021neural} and \cite{gong2023redeem}, the constant value $q$ is set to $2$.  
During the training of the evaluation framework, both the generator and the discriminator use the Adam optimizer for parameter optimization, with a learning rate set to 0.0003. The batch size for the MNIST dataset is set to 100, and for the AT\&T dataset, it is set to 40.

\textbf{Evaluation Metrics}. 
In the experiments, we assess the effectiveness and robustness of the proposed unlearning approach and evaluation framework.
Specifically, adhering to the validation strategy outlined in \cite{wu2022federated} and \cite{chundawat2023can}, we employ backdoor attack and membership inference attack to measure the unlearning capability. 
Furthermore, we use L2 distance and Jensen-Shannon Divergence (JSD) to quantify the similarity between the model after applying the proposed unlearning approach and the retrained model. A smaller JSD and L2 distance indicate a higher similarity between the two distributions.
 

\textbf{Baseline}. 
To compare with the most recent and relevant work and demonstrate the effectiveness of the proposed approach, we establish the following baselines.
The first baseline, denoted as $B_1$, retrains the model from scratch \cite{zhang2023fedrecovery}. The second baseline, $B_2$ \cite{chundawat2023can}, employs an incompetent teacher model similar to our unlearning approach, while the third baseline, $B_3$ \cite{li2021neural}, utilizes neural attention distillation, a technique also applied in our approach.

\subsection{Experimental Result}

\begin{figure*}[h]
\vspace{-0.42cm}
\centering
    \subfloat[]{\label{mnist-acc}\vspace{-0.1cm}\includegraphics[width=0.3\textwidth]{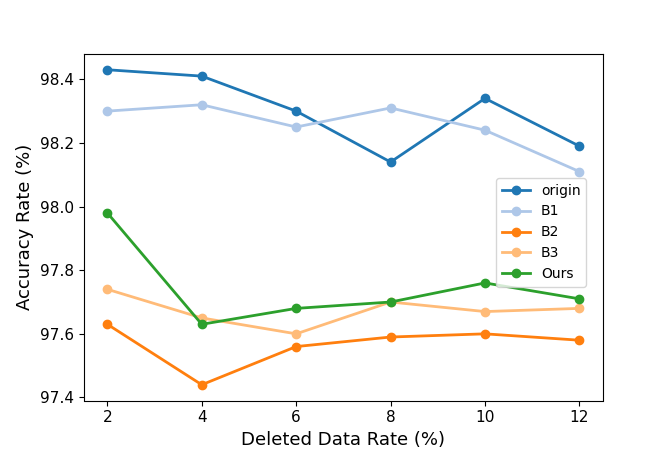}}
    \subfloat[]{\label{mnist-backdoor}\vspace{-0.1cm}\includegraphics[width=0.3\textwidth]{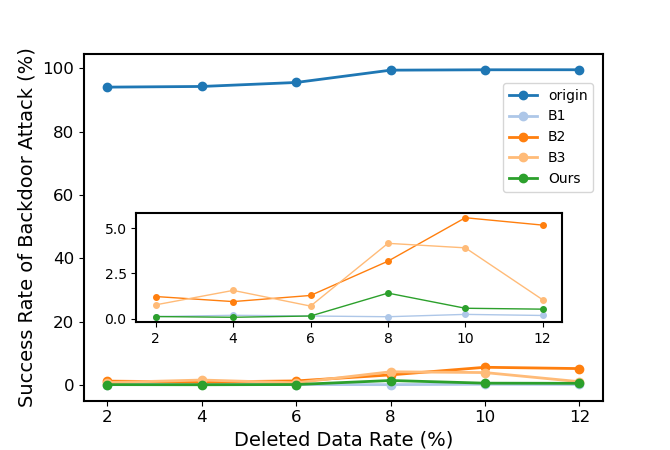}}
    \subfloat[]{\label{mnist-mem}\vspace{-0.1cm}\includegraphics[width=0.3\textwidth]{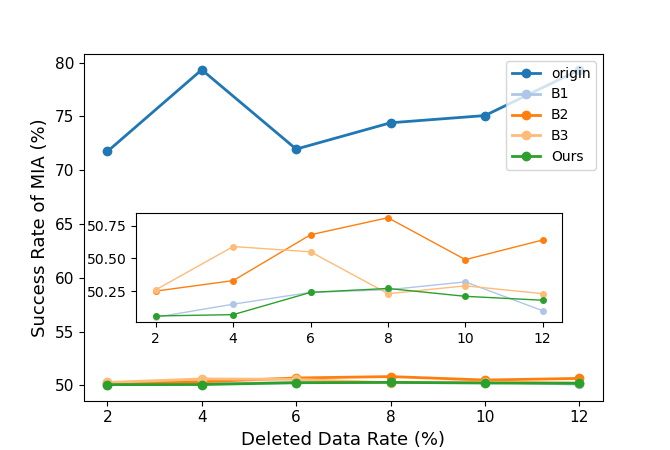}}
\vspace{-0.2cm}
 \caption{(a) Accuracy Rate, (b) Success Rate of Backdoor Attack, and (c) Success Rate of Membership Inference Attack of models on the MNIST dataset.}
 \label{fig:backdoor1}
\end{figure*} 
\begin{figure*}[h]
\centering
\vspace{-0.5cm}
    \subfloat[]{\label{cifar10-acc}\vspace{-0.1cm}\includegraphics[width=0.3\textwidth]{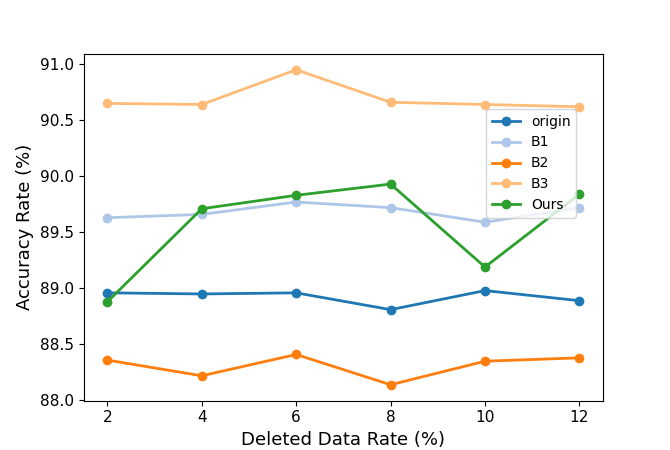}}
    \subfloat[]{\label{cifar10-backdoor}\vspace{-0.1cm}\includegraphics[width=0.3\textwidth]{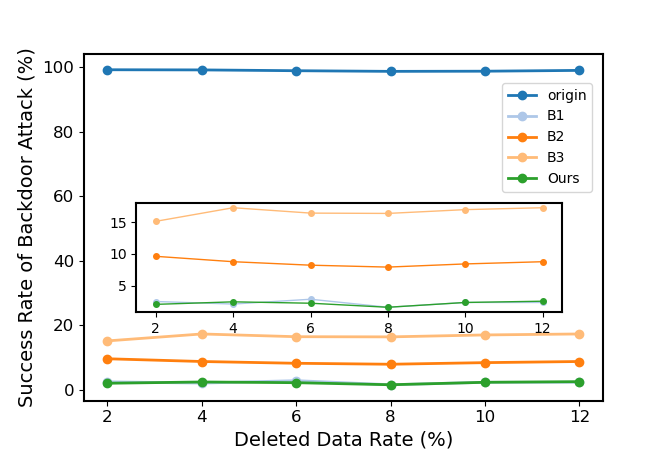}}
    \subfloat[]{\label{cifar10-mem}\vspace{-0.1cm}\includegraphics[width=0.3\textwidth]{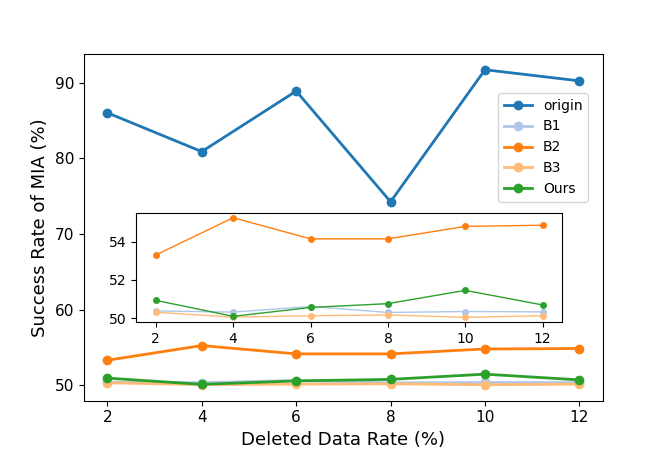}}
\vspace{-0.2cm}
 \caption{(a) Accuracy Rate, (b) Success Rate of Backdoor Attack, and (c) Success Rate of Membership Inference Attack of models on the CIFAR-10 dataset.}
 \label{fig:backdoor2}
\end{figure*} 
\begin{figure*}[!h]
\centering
\vspace{-0.5cm}
    \subfloat[]{\label{cifar100-acc}\vspace{-0.1cm}\includegraphics[width=0.3\textwidth]{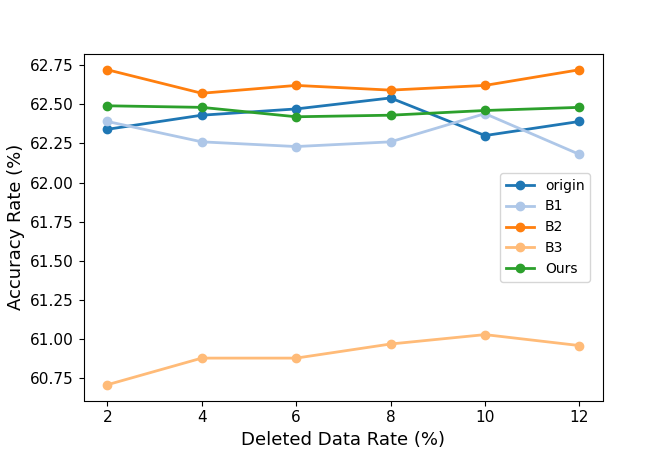}}
    \subfloat[]{\label{cifar100-backdoor}\vspace{-0.1cm}\includegraphics[width=0.3\textwidth]{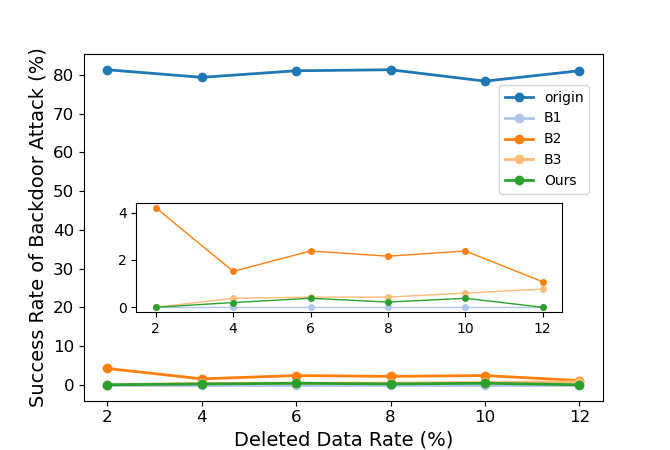}}
    \subfloat[]{\label{cifar100-mem}\vspace{-0.1cm}\includegraphics[width=0.3\textwidth]{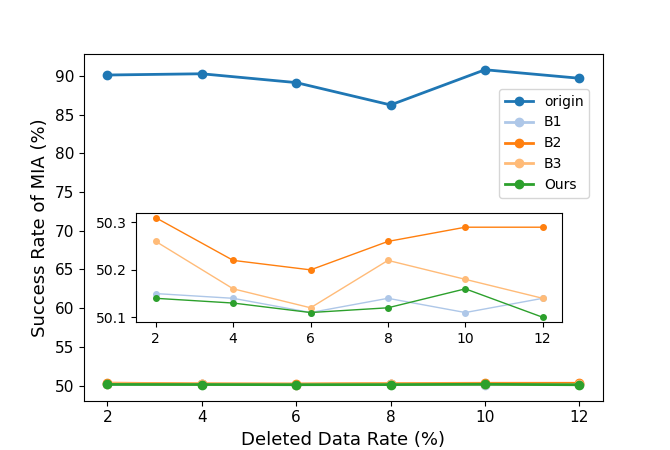}}
\vspace{-0.2cm}
 \caption{(a) Accuracy Rate, (b) Success Rate of Backdoor Attack, and (c) Success Rate of Membership Inference Attack of models on the CIFAR-100 dataset.}
 \label{fig:backdoor3}
\end{figure*} 

\subsubsection{Unlearning Approach Evaluation}

To comprehensively evaluate the proposed unlearning approach, we conduct experiments on sample unlearning, category unlearning, ablation studies, and improvement strategies.
Since most existing machine unlearning methods \cite{golatkar2020eternal,chen2023unlearn,kurmanji2024towards,jia2023model} target image classification scenarios  \cite{yao2024machine}, and our verification method prioritizes visualization, our experiments are conducted in the field of image processing.

\textbf{Sample unlearning}.
For sample unlearning, we adopt the method proposed by Wu \textit{et al.} \cite{wu2022federated},
 by altering the pixel values at certain specific locations of benign samples to embed a backdoor trigger,
where the proportion of backdoors corresponds to the data deletion rate. 
In our experiments, we use the same data deletion rates as those in Wang \textit{et al.} \cite{wang2023machine}.
After embedding the backdoors into the training dataset,  the model is trained. 
After that, we execute the proposed unlearning approach to unlearn the backdoor samples.
Finally, we measure the model's unlearning effect by the success rate of backdoor attacks.

In addition, we apply the membership inference attack \cite{chundawat2023can} to assess whether the deleted samples were part of the model's training process. A higher success rate in the membership inference attack suggests a less effective unlearning of the model on the deleted samples.
When the success rate approaches 50\%, it indicates that the model's unlearning effect is effective, as the attack method's confidence in successfully predicting the forgotten samples aligns closely with random prediction.

The experimental results for sample unlearning are shown in Fig.\ref{fig:backdoor1}, Fig.\ref{fig:backdoor2}, and Fig.\ref{fig:backdoor3}. 
From Fig.\ref{fig:backdoor1}, our experiments on the MNIST dataset show that $B_2$, $B_3$, and other methods experience a slight decrease in performance compared to the retrained and original models. However, our method achieves higher average accuracy among all baselines and is closer to $B_1$. Additionally, there is a significant reduction in both the success rate of backdoor attacks and membership inference attacks, often approaching the performance of $B_1$. This indicates that the unlearned model obtained by our method closely approximates the retrained model.
In Fig.\ref{fig:backdoor2}, based on experiments conducted on the CIFAR-10 dataset, our approach not only matches $B_1$ in accuracy but also achieves comparable results to $B_1$ in the success rate of backdoor attacks. Moreover, our approach demonstrates similar performance to $B_1$ and $B_3$ in terms of the success rate of membership inference attacks, which is significantly better than $B_2$ and the original model.
 In Fig.\ref{fig:backdoor3}, using the CIFAR-100 dataset, our approach achieves accuracy comparable to $B_1$ and better than $B_3$ but shows a slight decrease compared to $B_2$. However, our approach closely matches $B_1$ in terms of the success rate of backdoor attacks and membership inference attacks, outperforming the remaining methods. 
Based on the above results, we conclude that our proposed unlearning approach strikes a good balance between model accuracy and attack success rates. This demonstrates the effectiveness of our approach.

\begin{figure}[h]
\vspace{-0.3cm}
\centering
    \subfloat[]{\label{mnist-JSD}\vspace{-0.2cm}\includegraphics[width=0.24\textwidth]{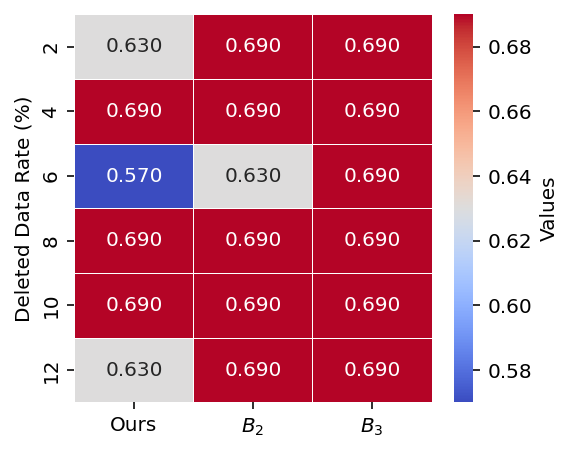}}
    \subfloat[]{\label{mnist-L2}\vspace{-0.2cm}\includegraphics[width=0.24\textwidth]{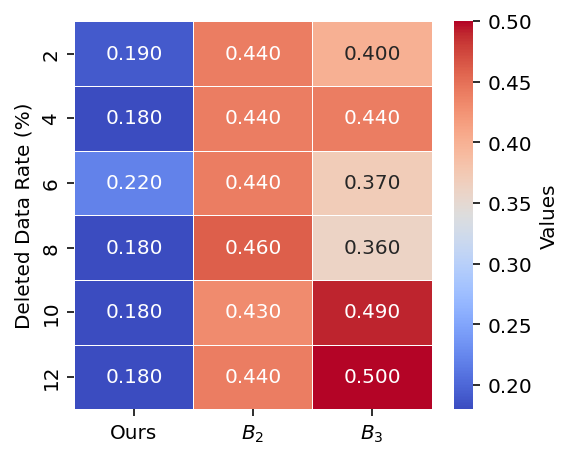}}
 \caption{Performance Metrics on MNIST dataset. (a) JSD, (b) L2.}
 \label{fig:mnist-metrics}
\end{figure} 

\begin{figure}[h]
\vspace{-0.3cm}
\centering
    \subfloat[]{\label{cifar10-JSD}\vspace{-0.2cm}\includegraphics[width=0.24\textwidth]{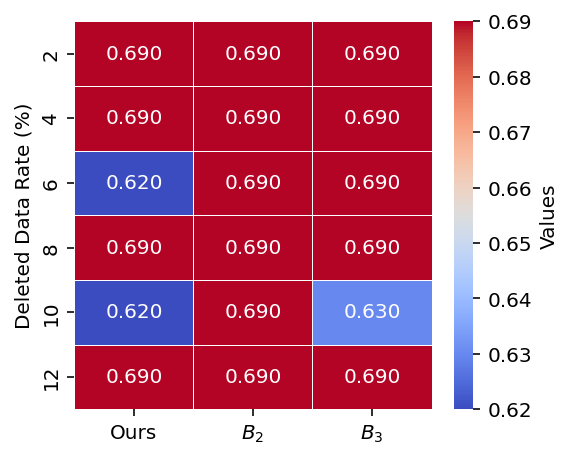}}
    \subfloat[]{\label{cifar10-L2}\vspace{-0.2cm}\includegraphics[width=0.24\textwidth]{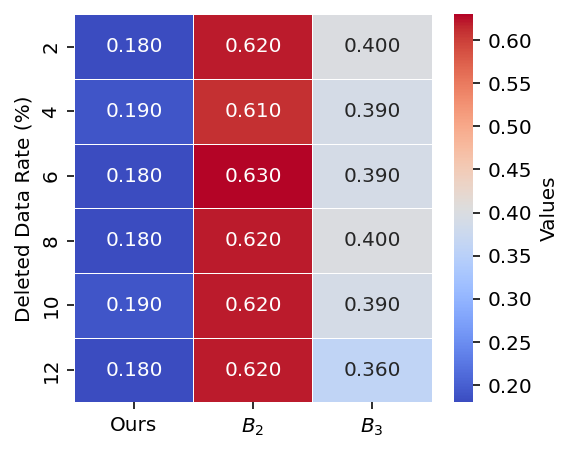}}
 \caption{Performance Metrics on CIFAR-10 dataset. (a) JSD, (b) L2.}
 \label{fig:cifar10-metrics}
\end{figure} 

\begin{figure}[h]
\vspace{-0.115cm}
\centering
    \subfloat[]{\label{cifar100-JSD}\vspace{-0.2cm}\includegraphics[width=0.2455\textwidth]{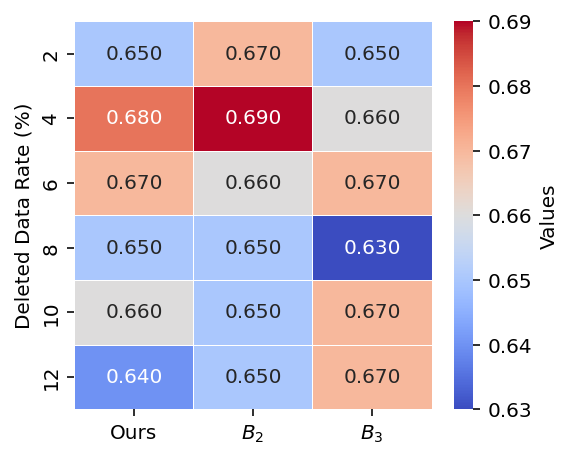}}
    \subfloat[]{\label{cifar100-L2}\vspace{-0.2cm}\includegraphics[width=0.24\textwidth]{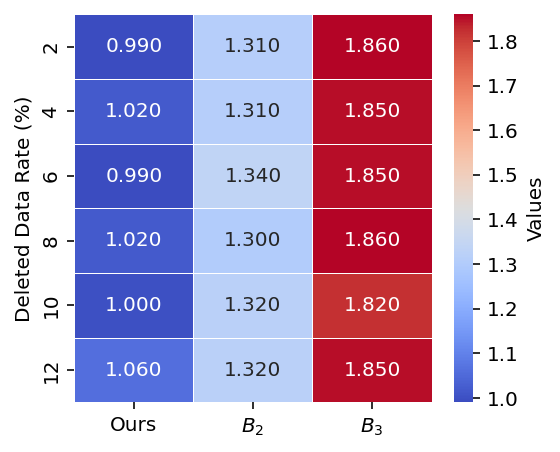}}
 \caption{Performance Metrics on CIFAR-100 dataset. (a) JSD, (b) L2.}
 \label{fig:cifar100-metrics}
\end{figure}

To further evaluate the effectiveness of the unlearning approach, we conduct experiments to assess the distribution similarity in model predictions between the retrained model and other models.
The results are shown in Fig.\ref{fig:mnist-metrics}, Fig.\ref{fig:cifar10-metrics}, and Fig.\ref{fig:cifar100-metrics}.
We can observe that across all data deletion rates (2\%, 4\%, 6\%, 8\%, 10\%, 12\%), our approach exhibits smaller values in both L2 distance and JSD divergence, demonstrating that our proposed approach is closer to $B_1$ in terms of predictive patterns.

\textbf{Category unlearning}.
For the category unlearning task, we follow the methodology outlined in \cite{chundawat2023can}, using the MNIST, CIFAR-10, and CIFAR-100 datasets to study category unlearning. Specifically, we focus on single-category unlearning and randomly select 20\% of the total categories for unlearning. We measure the model's test accuracy separately for the remaining categories and the unlearned categories.
We use evaluation metrics to assess the unlearning performance of both baselines and our approach. The experimental results are summarized in  Table \ref{Category forgetting}.
 From Table \ref{Category forgetting}, our approach demonstrates superior performance compared to baselines $B_2$ and $B_3$ in deleted data. Moreover, our approach achieves comparable performance to $B_1$ on the remaining data.

\begin{table*}[t]
\centering 
\tabcolsep=0.4cm
\caption{Category unlearning on MNIST, CIFAR-10, and CIFAR-100 datasets.}
\label{Category forgetting}
\renewcommand\arraystretch{1.1}
\begin{tabular}{cccccccc}
\hline
Dataset                   & Unlearn Class number & Metrics      & origin  & $B_1$    & $B_2$    & $B_3$    & Ours    \\ \hline
\multirow{4}{*}{MNIST}    & \multirow{2}{*}{1}   & acc on $D_r$$\uparrow$ & 98.24\% & 98.64\% & 97.42\% & 96.54\% & 97.54\% \\
                          &                      & acc on $D_f$$\downarrow$ & 99.74\% & 0.00\%  & 0.09\%  & 29.16\% & 0.00\%  \\
                          & \multirow{2}{*}{2}   & acc on $D_r$$\uparrow$ & 98.36\% & 98.94\% & 98.29\% & 95.75\% & 97.90\% \\
                          &                      & acc on $D_f$$\downarrow$ & 98.85\% & 0.00\%  & 0.47\%  & 27.13\% & 0.00\%  \\ \hline
\multirow{4}{*}{CIFAR-10}  & \multirow{2}{*}{1}   & acc on $D_r$$\uparrow$ & 86.01\% & 78.90\% & 77.30\% & 76.30\% & 78.78\% \\
                          &                      & acc on $D_f$$\downarrow$ & 94.80\% & 0.00\%  & 4.60\%  & 24.20\% & 0.00\%  \\
                          & \multirow{2}{*}{2}   & acc on $D_r$$\uparrow$ & 86.68\% & 78.65\% & 77.20\% & 75.23\% & 78.84\% \\
                          &                      & acc on $D_f$$\downarrow$ & 87.75\% & 0.00\%  & 8.59\%  & 22.20\% & 0.00\%  \\ \hline
\multirow{4}{*}{CIFAR-100} & \multirow{2}{*}{1}   & acc on $D_r$$\uparrow$ & 61.81\% & 61.35\% & 62.16\% & 57.29\% & 61.30\% \\
                          &                      & acc on $D_f$$\downarrow$ & 70.00\% & 0.00\%  & 0.00\%  & 7.00\%  & 0.00\%  \\
                          & \multirow{2}{*}{20}  & acc on $D_r$$\uparrow$ & 61.81\% & 59.61\% & 57.91\% & 53.17\% & 58.61\% \\
                          &                      & acc on $D_f$$\downarrow$ & 60.65\% & 0.00\%  & 17.13\% & 30.95\% & 15.38\% \\ \hline
\end{tabular}
\end{table*}

\begin{table*}[]
\centering 
\caption{The attention map discrepancies calculated on MNIST, CIFAR-10, and CIFAR-100 datasets.}
\label{AMA}
\tabcolsep=0.35cm
\renewcommand\arraystretch{1.1}
\begin{tabular}{ccccccccc}
\hline
\multirow{2}{*}{Dataset} & \multicolumn{2}{c}{original}                                                                                         & \multicolumn{2}{c}{$B2$}                                                                                             & \multicolumn{2}{c}{$B3$}                                                                                             & \multicolumn{2}{c}{Ours}                                                                                             \\ \cline{2-9} 
                         & \begin{tabular}[c]{@{}c@{}}backdoored\\ sample\end{tabular} & \begin{tabular}[c]{@{}c@{}}clean\\ sample\end{tabular} & \begin{tabular}[c]{@{}c@{}}backdoored\\ sample\end{tabular} & \begin{tabular}[c]{@{}c@{}}clean\\ sample\end{tabular} & \begin{tabular}[c]{@{}c@{}}backdoored\\ sample\end{tabular} & \begin{tabular}[c]{@{}c@{}}clean\\ sample\end{tabular} & \begin{tabular}[c]{@{}c@{}}backdoored\\ sample\end{tabular} & \begin{tabular}[c]{@{}c@{}}clean\\ sample\end{tabular} \\ \hline
MNIST                    & 4.57                                                        & 4.59                                                   & 13.45                                                       & 15.65                                                  & 2.57                                                        & 2.65                                                   & 1.53                                                        & 1.56                                                   \\
CIFAR-10                 & 6.63                                                        & 6.49                                                   & 5.42                                                        & 5.80                                                   & 2.98                                                        & 3.66                                                   & 1.78                                                        & 1.72                                                   \\
CIFAR-100                & 3.59                                                        & 3.68                                                   & 2.78                                                        & 3.15                                                   & 2.66                                                        & 2.99                                                   & 2.36                                                        & 2.46                                                   \\ \hline
\end{tabular}
\end{table*}

\begin{figure}[t]
    \centering
    \includegraphics[width=1\linewidth]{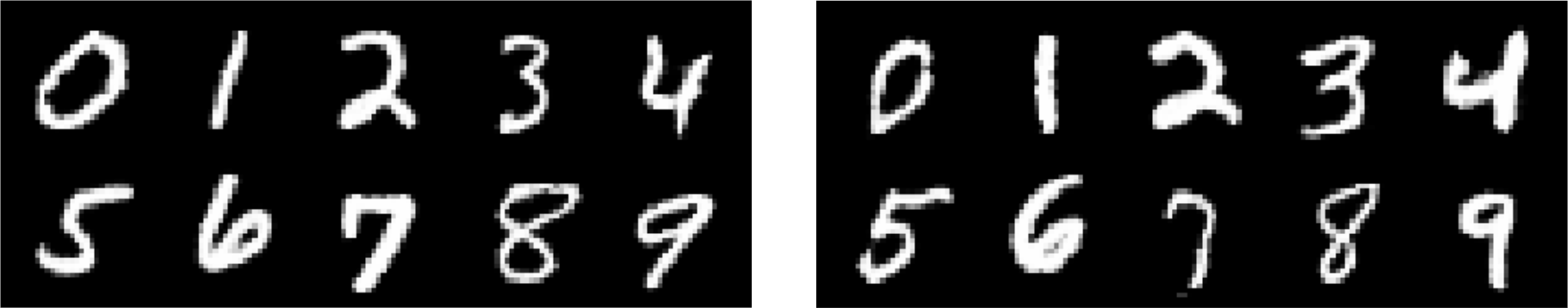}
    \caption{Samples Generated by Skyeye with a competent Classifier trained on MNIST dataset.}
    \label{Samples Generated by Evaluation Framework with a competent Classifier trained on MNIST dataset.}
\end{figure}
\begin{figure}[]
    \centering
    \includegraphics[width=1\linewidth]{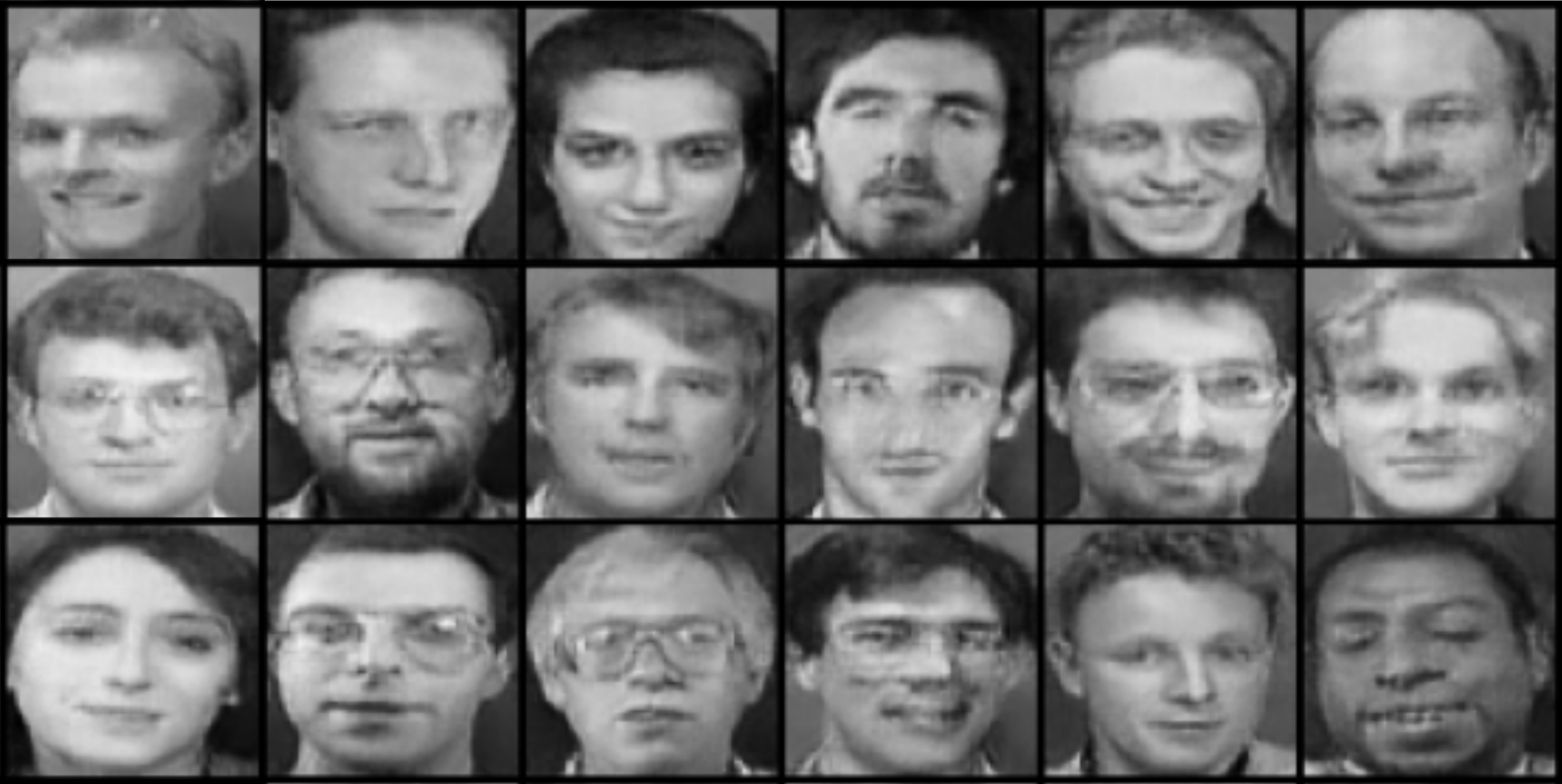}
    \caption{Samples Generated by Skyeye with a competent Classifier trained on AT\&T dataset.}
    \label{Samples Generated by Evaluation Framework with a competent Classifier trained on AT dataset.}
\end{figure}
\begin{figure}[t]
    \centering
    \includegraphics[width=1\linewidth]{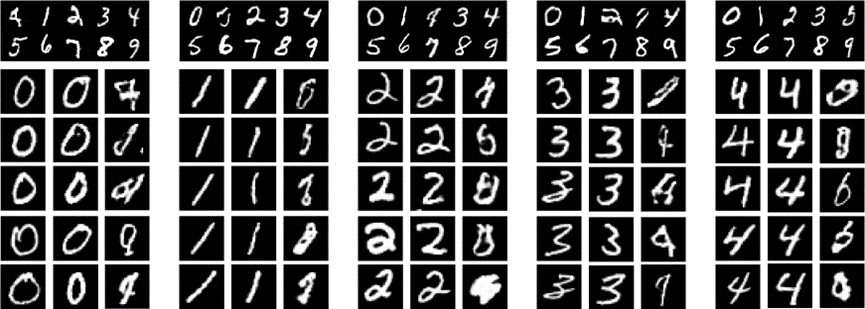}
    \caption{Samples Generated by Skyeye with an unlearned Classifier trained on MNIST dataset.
    From left to right, the images displayed are samples generated by the classifier following the unlearning of the digits 0, 1, 2, 3, and 4. 
    For each group, the top part provides an overview of all the categories generated and the subsequent rows correspond to the forgotten class, the image on the left is the real image, the middle is the one generated by the generator before unlearning, and the image on the right is the one produced by the generator after unlearning.}
    \label{Samples Generated by Evaluation Framework with an unlearned Classifier trained on MNIST dataset.}
\end{figure}

\textbf{The effectiveness of Attention Map Alignment.}
To verify the effectiveness of Attention Map Alignment in our federated unlearning approach, we compare the differences between the attention maps generated by various unlearning methods and those obtained from the retrained model.
The dataset used includes both backdoored and clean samples.
Table \ref{AMA} demonstrates that our approach results in the smallest attention map discrepancies compared to all other methods.
Combining the results from Fig.\ref{Category forgetting}, it is evident that lower attention map discrepancies lead to better performance in both accuracy and robustness, thereby demonstrating the effectiveness of attention map alignment techniques.


\textbf{Ablation study of loss function.}
To investigate the significance of different components of the loss function, we conducted an ablation study on the loss function. 
We refer the reader to Appendix \customref{D} for further details.

\begin{figure}[]
    \centering
    \includegraphics[width=1\linewidth]{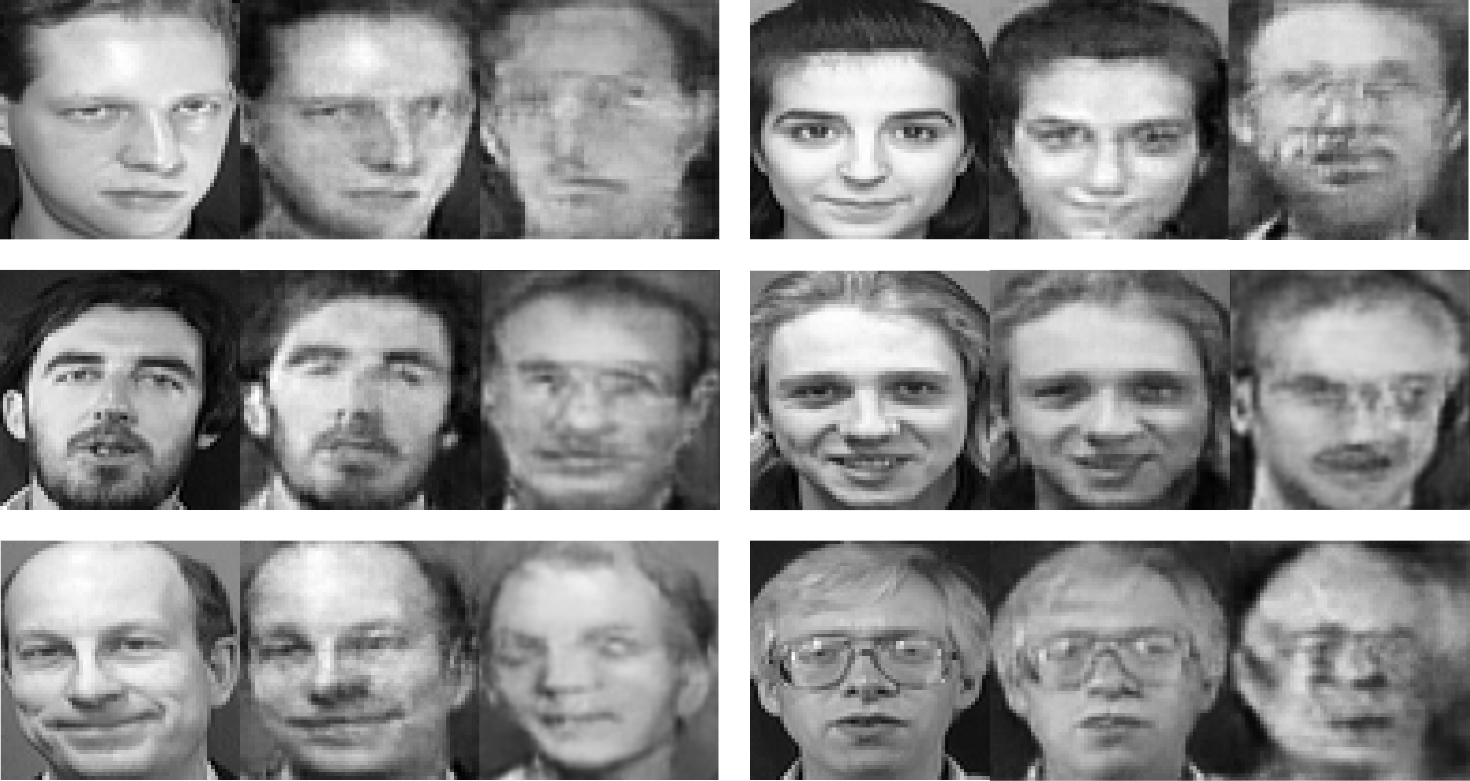}
    \caption{Samples Generated by Skyeye with an unlearned Classifier trained on AT\&T dataset.
    There are six sets of sample images, each consisting of three images:
    the left image is the real image from the dataset; the middle image is generated by the generator before unlearning; The right image is generated by the generator after unlearning.
    }
    \label{Samples Generated by Evaluation Framework with an unlearned Classifier trained on AT&T dataset.}
\end{figure}
\begin{figure}[t]
    \centering
    \includegraphics[width=1\linewidth]{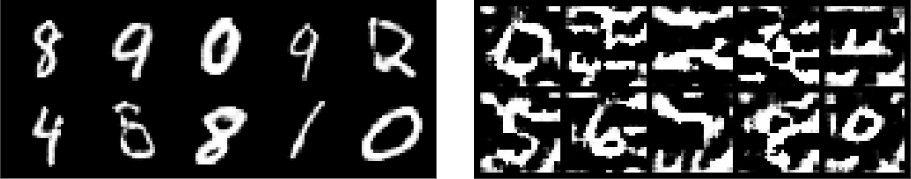}
    \caption{The ablation study on the generator's loss on MNIST dataset.
    The image on the left shows samples generated by the generator trained with only the discriminator loss; the image on the right shows samples generated by the generator trained with only the classifier loss.
}
    \label{The Ablation studies on the generator's loss on MNIST.}
\end{figure}

\begin{figure}[t]
    \centering
    \includegraphics[width=1\linewidth]{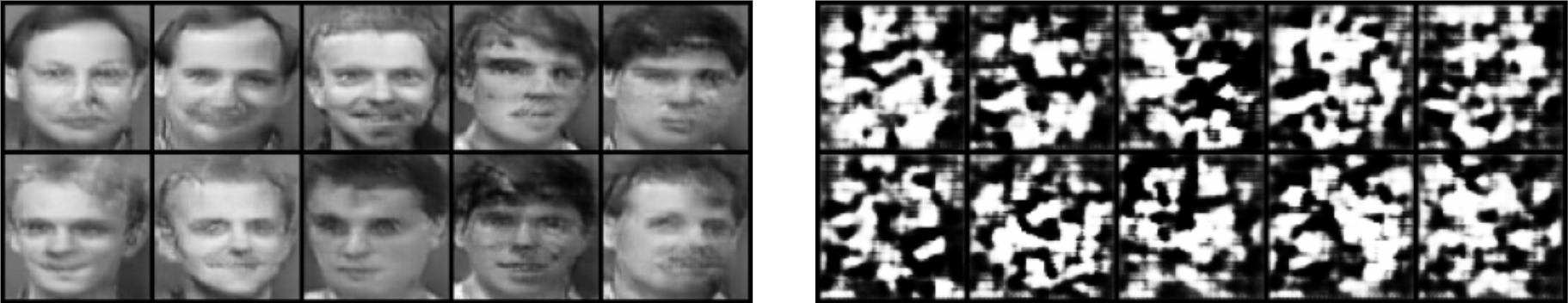}
    \caption{The ablation study on the generator's loss on AT\&T dataset.
   The image on the left shows samples generated by the generator trained with only the discriminator loss; the image on the right shows samples generated by the generator trained with only the classifier loss.}
    \label{The Ablation studies on the generator's loss on AT&T.}
\end{figure}

\begin{table*}[!ht]
\centering 
\tabcolsep=0.275cm
\caption{Ablation Study of the hyperparameters of the mechanisms proposed in our Further Improvements.}
\label{hyperparameters of the mechanisms}
\renewcommand\arraystretch{1.1}
\begin{tabular}{ccccccc}
\hline
Learning Rate Decay & $1.0\times 10^{-1}$ & $5.0\times 10^{-2}$ & 1.0$\times 10^{-2}$ & $5.0\times 10^{-3}$ & $1.0\times 10^{-3}$ & $5.0\times 10^{-4}$ \\ \hline
backdoor$\downarrow$                & 0.00\%  & 0.00\%  & 1.09\% & 2.44\%  & 28.23\%  & 77.53\% \\
acc$\uparrow$                     & 33.15\% & 66.11\% & 81.93\% & 85.44\% & 86.09\% & 88.68\% \\ \hline
Bounded Loss           & 0       & 5       & 10      & 15      & 20      & 25      \\ \hline
backdoor$\downarrow$                & 0.00\%  & 74.37\% & 37.66\% & 2.34\%  & 0.00\%  & 0.00\%  \\
acc$\uparrow$                     & 11.71\% & 87.72\% & 84.48\% & 83.50\% & 75.84\% & 73.33\% \\ \hline
Preventing Catastrophic Forgetting & 0       & 0.2     & 0.4     & 0.6     & 0.8     & 1       \\ \hline
backdoor$\downarrow$                & 0.00\%  & 0.00\%  & 1.32\%  & 8.54\%  & 17.41\% & 76.46\% \\
acc$\uparrow$                     & 74.77\% & 76.58\% & 79.27\% & 79.64\% & 80.12\% & 82.15\%   \\ \hline
\end{tabular}
\end{table*}


\textbf{Improvement strategy evaluation}.
To evaluate the effectiveness of the mechanisms proposed in our improvement module for preventing catastrophic forgetting, bounded loss, and learning rate decay, we conducted an ablation study on the hyperparameters. 
The experimental results are shown in Table \ref{hyperparameters of the mechanisms}. 
From the table, we observe that the Learning Rate Decay parameter adjusts the size of the learning rate throughout the unlearning process. 
A larger learning rate can lead to excessive unlearning, potentially damaging the model's accuracy, whereas a smaller learning rate may result in ineffective unlearning. Setting the learning rate to 
 \( 5.0 \times 10^{-3} \) enables the model to effectively converge to an appropriate point, balancing between unlearning unwanted information and maintaining model accuracy.

Bounded Loss adjusts the maximum allowable increase in the absolute value of the hard loss. 
If the $BND$ is equal to $0$, the bounded loss mechanism is not utilized.
A smaller $BND$ value can help maintain higher model accuracy but may fail to reduce the success rate of a backdoor attack.
A larger $BND$ value also harms the model's accuracy, potentially causing an over-unlearning situation.
When $BND$ is set to 15, the model maintains high accuracy and achieves a backdoor attack success rate comparable to the retrained model.

To prevent catastrophic forgetting, it is crucial to adjust the weight of the regularization term that controls the deviation of model parameters.
When the weight of the regularization term is small, the regularization strength is insufficient, resulting in a decrease in the success rate of the backdoor attack while also causing a significant decline in model accuracy. 
Conversely, when the weight is large, it restricts the model's convergence range, which may lead to the retention of the backdoor pattern. 
When the weight is set to $0.4$, a balance is struck between maintaining model accuracy and reducing the success rate of the backdoor attack.

\subsubsection{Skyeye Framework Evaluation}
Our Skyeye framework is the first visualization assessment framework 
 primarily designed to visually evaluate the capability of category unlearning.
It relies on the assumption that if a model has successfully unlearned a category, the generator should no longer be able to effectively produce images of that category.

To set up Skyeye, we begin by training classifiers on target datasets. 
Specifically, we develop separate classifiers tailored for the MNIST and AT\&T datasets. 
After training the classifiers on the target datasets, we integrate them with the applied GAN network.
We then utilize the classifier's knowledge to guide the generator in producing images.
For example, Fig.\ref{Samples Generated by Evaluation Framework with a competent Classifier trained on MNIST dataset.} illustrates the knowledge of the classier corresponding to the MNIST  while Fig.\ref{Samples Generated by Evaluation Framework with a competent Classifier trained on AT dataset.} demonstrates the knowledge of the classier corresponding to the AT\&T datasets respectively.

It is evident that for both datasets, the generator is capable of accurately producing samples corresponding to the respective categories. This observation demonstrates that our evaluation method effectively visualizes the implicit knowledge learned by the classifier through the generator. 



The classifier employs the $B_1$ approach to unlearn a specific category. We then observe the differences in the images generated by the generator before and after unlearning. 
The experimental results are shown in Fig.\ref{Samples Generated by Evaluation Framework with an unlearned Classifier trained on MNIST dataset.} and Fig.\ref{Samples Generated by Evaluation Framework with an unlearned Classifier trained on AT&T dataset.}.
From the results, we observe that the generator successfully generates images for all categories except the unlearned category.

Next, we conduct an ablation study on the generator's loss function, specifically examining the effects of discriminator loss and classification loss. 
The results of the experiment are shown in Fig.\ref{The Ablation studies on the generator's loss on MNIST.} and Fig.\ref{The Ablation studies on the generator's loss on AT&T.}.
From the results, we can infer that without the discriminator, the generator is unable to properly express the implicit knowledge possessed by the classifier.

\section{Conclusion}
In this paper, we propose a federated unlearning approach that effectively unlearns by improving the loss function and adding further improvements. 
At the same time, we introduce a new paradigm to evaluate the effectiveness of federated unlearning algorithms by leveraging the data recovery capabilities of GANs. Furthermore, we propose an unlearning evaluation framework called Skyeye, which uses a GAN network to visualize the internal knowledge in the unlearned model and judge the effectiveness of unlearning by observing the generated content.
Furthermore, we conduct comprehensive experiments to validate the effectiveness of both the unlearning approach and Skyeye framework.

\bibliographystyle{ieeetr}
\bibliography{reference}

@article{jordan2015machine,
  title={Machine learning: Trends, perspectives, and prospects},
  author={Jordan, Michael I and Mitchell, Tom M},
  journal={Science},
  volume={349},
  number={6245},
  pages={255--260},
  year={2015},
  publisher={American Association for the Advancement of Science}
}

@article{pardau2018california,
  title={The california consumer privacy act: Towards a european-style privacy regime in the united states},
  author={Pardau, Stuart L},
  journal={J. Tech. L. \& Pol'y},
  volume={23},
  pages={68},
  year={2018},
  publisher={HeinOnline}
}

@inproceedings{chen2021machine,
  title={When machine unlearning jeopardizes privacy},
  author={Chen, Min and Zhang, Zhikun and Wang, Tianhao and Backes, Michael and Humbert, Mathias and Zhang, Yang},
  booktitle={Proceedings of the 2021 ACM SIGSAC conference on computer and communications security},
  pages={896--911},
  year={2021}
}

@inproceedings{cao2015towards,
  title={Towards making systems forget with machine unlearning},
  author={Cao, Yinzhi and Yang, Junfeng},
  booktitle={2015 IEEE symposium on security and privacy},
  pages={463--480},
  year={2015},
  organization={IEEE}
}

@article{xu2024machine,
  title={Machine unlearning: Solutions and challenges},
  author={Xu, Jie and Wu, Zihan and Wang, Cong and Jia, Xiaohua},
  journal={IEEE Transactions on Emerging Topics in Computational Intelligence},
  year={2024},
  publisher={IEEE}
}

@inproceedings{bourtoule2021machine,
  title={Machine unlearning},
  author={Bourtoule, Lucas and Chandrasekaran, Varun and Choquette-Choo, Christopher A and Jia, Hengrui and Travers, Adelin and Zhang, Baiwu and Lie, David and Papernot, Nicolas},
  booktitle={2021 IEEE Symposium on Security and Privacy (SP)},
  pages={141--159},
  year={2021},
  organization={IEEE}
}

@inproceedings{brophy2021machine,
  title={Machine unlearning for random forests},
  author={Brophy, Jonathan and Lowd, Daniel},
  booktitle={International Conference on Machine Learning},
  pages={1092--1104},
  year={2021},
  organization={PMLR}
}

@article{ginart2019making,
  title={Making ai forget you: Data deletion in machine learning},
  author={Ginart, Antonio and Guan, Melody and Valiant, Gregory and Zou, James Y},
  journal={Advances in neural information processing systems},
  volume={32},
  year={2019}
}

@inproceedings{su2023asynchronous,
  title={Asynchronous federated unlearning},
  author={Su, Ningxin and Li, Baochun},
  booktitle={IEEE INFOCOM 2023-IEEE Conference on Computer Communications},
  pages={1--10},
  year={2023},
  organization={IEEE}
}

@inproceedings{liu2022right,
  title={The right to be forgotten in federated learning: An efficient realization with rapid retraining},
  author={Liu, Yi and Xu, Lei and Yuan, Xingliang and Wang, Cong and Li, Bo},
  booktitle={IEEE INFOCOM 2022-IEEE Conference on Computer Communications},
  pages={1749--1758},
  year={2022},
  organization={IEEE}
}

@article{guo2019certified,
  title={Certified data removal from machine learning models},
  author={Guo, Chuan and Goldstein, Tom and Hannun, Awni and Van Der Maaten, Laurens},
  journal={arXiv preprint arXiv:1911.03030},
  year={2019}
}

@article{sekhari2021remember,
  title={Remember what you want to forget: Algorithms for machine unlearning},
  author={Sekhari, Ayush and Acharya, Jayadev and Kamath, Gautam and Suresh, Ananda Theertha},
  journal={Advances in Neural Information Processing Systems},
  volume={34},
  pages={18075--18086},
  year={2021}
}

@inproceedings{golatkar2020eternal,
  title={Eternal sunshine of the spotless net: Selective forgetting in deep networks},
  author={Golatkar, Aditya and Achille, Alessandro and Soatto, Stefano},
  booktitle={Proceedings of the IEEE/CVF Conference on Computer Vision and Pattern Recognition},
  pages={9304--9312},
  year={2020}
}

@inproceedings{wu2020deltagrad,
  title={Deltagrad: Rapid retraining of machine learning models},
  author={Wu, Yinjun and Dobriban, Edgar and Davidson, Susan},
  booktitle={International Conference on Machine Learning},
  pages={10355--10366},
  year={2020},
  organization={PMLR}
}

@inproceedings{mcmahan2017communication,
  title={Communication-efficient learning of deep networks from decentralized data},
  author={McMahan, Brendan and Moore, Eider and Ramage, Daniel and Hampson, Seth and y Arcas, Blaise Aguera},
  booktitle={Artificial intelligence and statistics},
  pages={1273--1282},
  year={2017},
  organization={PMLR}
}

@article{konevcny2016federated,
  title={Federated learning: Strategies for improving communication efficiency},
  author={Kone{\v{c}}n{\`y}, Jakub and McMahan, H Brendan and Yu, Felix X and Richt{\'a}rik, Peter and Suresh, Ananda Theertha and Bacon, Dave},
  journal={arXiv preprint arXiv:1610.05492},
  year={2016}
}

@article{regulation2018general,
  title={General data protection regulation (GDPR)},
  author={Regulation, General Data Protection},
  journal={Intersoft Consulting, Accessed in October},
  volume={24},
  number={1},
  year={2018}
}

@article{salem2018ml,
  title={Ml-leaks: Model and data independent membership inference attacks and defenses on machine learning models},
  author={Salem, Ahmed and Zhang, Yang and Humbert, Mathias and Berrang, Pascal and Fritz, Mario and Backes, Michael},
  journal={arXiv preprint arXiv:1806.01246},
  year={2018}
}

@inproceedings{bagdasaryan2020backdoor,
  title={How to backdoor federated learning},
  author={Bagdasaryan, Eugene and Veit, Andreas and Hua, Yiqing and Estrin, Deborah and Shmatikov, Vitaly},
  booktitle={International conference on artificial intelligence and statistics},
  pages={2938--2948},
  year={2020},
  organization={PMLR}
}

@inproceedings{izzo2021approximate,
  title={Approximate data deletion from machine learning models},
  author={Izzo, Zachary and Smart, Mary Anne and Chaudhuri, Kamalika and Zou, James},
  booktitle={International Conference on Artificial Intelligence and Statistics},
  pages={2008--2016},
  year={2021},
  organization={PMLR}
}

@inproceedings{neel2021descent,
  title={Descent-to-delete: Gradient-based methods for machine unlearning},
  author={Neel, Seth and Roth, Aaron and Sharifi-Malvajerdi, Saeed},
  booktitle={Algorithmic Learning Theory},
  pages={931--962},
  year={2021},
  organization={PMLR}
}

@inproceedings{ullah2021machine,
  title={Machine unlearning via algorithmic stability},
  author={Ullah, Enayat and Mai, Tung and Rao, Anup and Rossi, Ryan A and Arora, Raman},
  booktitle={Conference on Learning Theory},
  pages={4126--4142},
  year={2021},
  organization={PMLR}
}

@inproceedings{nasr2018comprehensive,
  title={Comprehensive privacy analysis of deep learning},
  author={Nasr, Milad and Shokri, Reza and Houmansadr, Amir},
  booktitle={Proceedings of the 2019 IEEE Symposium on Security and Privacy (SP)},
  pages={1--15},
  year={2018}
}

@article{song2020analyzing,
  title={Analyzing user-level privacy attack against federated learning},
  author={Song, Mengkai and Wang, Zhibo and Zhang, Zhifei and Song, Yang and Wang, Qian and Ren, Ju and Qi, Hairong},
  journal={IEEE Journal on Selected Areas in Communications},
  volume={38},
  number={10},
  pages={2430--2444},
  year={2020},
  publisher={IEEE}
}

@inproceedings{truedata ,
  title={Forget unlearning: Towards true data-deletion in machine learning},
  author={Chourasia, Rishav and Shah, Neil},
  booktitle={International Conference on Machine Learning},
  pages={6028--6073},
  year={2023},
  organization={PMLR}
}

@article{gupta2021adaptive,
  title={Adaptive machine unlearning},
  author={Gupta, Varun and Jung, Christopher and Neel, Seth and Roth, Aaron and Sharifi-Malvajerdi, Saeed and Waites, Chris},
  journal={Advances in Neural Information Processing Systems},
  volume={34},
  pages={16319--16330},
  year={2021}
}

@article{baumhauer2022machine,
  title={Machine unlearning: Linear filtration for logit-based classifiers},
  author={Baumhauer, Thomas and Sch{\"o}ttle, Pascal and Zeppelzauer, Matthias},
  journal={Machine Learning},
  volume={111},
  number={9},
  pages={3203--3226},
  year={2022},
  publisher={Springer}
}

@article{zhang2023fedrecovery,
  title={Fedrecovery: Differentially private machine unlearning for federated learning frameworks},
  author={Zhang, Lefeng and Zhu, Tianqing and Zhang, Haibin and Xiong, Ping and Zhou, Wanlei},
  journal={IEEE Transactions on Information Forensics and Security},
  year={2023},
  publisher={IEEE}
}

@inproceedings{liu2021federaser,
  title={Federaser: Enabling efficient client-level data removal from federated learning models},
  author={Liu, Gaoyang and Ma, Xiaoqiang and Yang, Yang and Wang, Chen and Liu, Jiangchuan},
  booktitle={2021 IEEE/ACM 29th International Symposium on Quality of Service (IWQOS)},
  pages={1--10},
  year={2021},
  organization={IEEE}
}

@inproceedings{thudi2022necessity,
  title={On the necessity of auditable algorithmic definitions for machine unlearning},
  author={Thudi, Anvith and Jia, Hengrui and Shumailov, Ilia and Papernot, Nicolas},
  booktitle={31st USENIX Security Symposium (USENIX Security 22)},
  pages={4007--4022},
  year={2022}
}

@inproceedings{chundawat2023can,
  title={Can bad teaching induce forgetting? Unlearning in deep networks using an incompetent teacher},
  author={Chundawat, Vikram S and Tarun, Ayush K and Mandal, Murari and Kankanhalli, Mohan},
  booktitle={Proceedings of the AAAI Conference on Artificial Intelligence},
  volume={37},
  number={6},
  pages={7210--7217},
  year={2023}
}

@inproceedings{chourasia2023forget,
  title={Forget unlearning: Towards true data-deletion in machine learning},
  author={Chourasia, Rishav and Shah, Neil},
  booktitle={International Conference on Machine Learning},
  pages={6028--6073},
  year={2023},
  organization={PMLR}
}

@article{lecun1998gradient,
  title={Gradient-based learning applied to document recognition},
  author={LeCun, Yann and Bottou, L{\'e}on and Bengio, Yoshua and Haffner, Patrick},
  journal={Proceedings of the IEEE},
  volume={86},
  number={11},
  pages={2278--2324},
  year={1998},
  publisher={Ieee}
}

@article{krizhevsky2009learning,
  title={Learning multiple layers of features from tiny images},
  author={Krizhevsky, Alex and Hinton, Geoffrey and others},
  year={2009},
  publisher={Toronto, ON, Canada}
}

@article{wang2023machine,
  title={Machine Unlearning via Representation Forgetting with Parameter Self-Sharing},
  author={Wang, Weiqi and Zhang, Chenhan and Tian, Zhiyi and Yu, Shui},
  journal={IEEE Transactions on Information Forensics and Security},
  year={2023},
  publisher={IEEE}
}

@article{wu2022federated,
  title={Federated unlearning with knowledge distillation},
  author={Wu, Chen and Zhu, Sencun and Mitra, Prasenjit},
  journal={arXiv preprint arXiv:2201.09441},
  year={2022}
}

@inproceedings{he2016deep,
  title={Deep residual learning for image recognition},
  author={He, Kaiming and Zhang, Xiangyu and Ren, Shaoqing and Sun, Jian},
  booktitle={Proceedings of the IEEE conference on computer vision and pattern recognition},
  pages={770--778},
  year={2016}
}

@article{shumailov2021manipulating,
  title={Manipulating sgd with data ordering attacks},
  author={Shumailov, Ilia and Shumaylov, Zakhar and Kazhdan, Dmitry and Zhao, Yiren and Papernot, Nicolas and Erdogdu, Murat A and Anderson, Ross J},
  journal={Advances in Neural Information Processing Systems},
  volume={34},
  pages={18021--18032},
  year={2021}
}

@article{gao2024verifi,
  title={Verifi: Towards verifiable federated unlearning},
  author={Gao, Xiangshan and Ma, Xingjun and Wang, Jingyi and Sun, Youcheng and Li, Bo and Ji, Shouling and Cheng, Peng and Chen, Jiming},
  journal={IEEE Transactions on Dependable and Secure Computing},
  year={2024},
  publisher={IEEE}
}

@article{sommer2022athena,
  title={Athena: Probabilistic verification of machine unlearning},
  author={Sommer, David M and Song, Liwei and Wagh, Sameer and Mittal, Prateek},
  journal={Proceedings on Privacy Enhancing Technologies},
  year={2022}
}

@article{guo2023verifying,
  title={Verifying in the Dark: Verifiable Machine Unlearning by Using Invisible Backdoor Triggers},
  author={Guo, Yu and Zhao, Yu and Hou, Saihui and Wang, Cong and Jia, Xiaohua},
  journal={IEEE Transactions on Information Forensics and Security},
  year={2023},
  publisher={IEEE}
}

@inproceedings{yue2023gradient,
  title={Gradient obfuscation gives a false sense of security in federated learning},
  author={Yue, Kai and Jin, Richeng and Wong, Chau-Wai and Baron, Dror and Dai, Huaiyu},
  booktitle={32nd USENIX Security Symposium (USENIX Security 23)},
  pages={6381--6398},
  year={2023}
}

@inproceedings{hitaj2017deep,
  title={Deep models under the GAN: information leakage from collaborative deep learning},
  author={Hitaj, Briland and Ateniese, Giuseppe and Perez-Cruz, Fernando},
  booktitle={Proceedings of the 2017 ACM SIGSAC conference on computer and communications security},
  pages={603--618},
  year={2017}
}

@article{goodfellow2014generative,
  title={Generative adversarial nets},
  author={Goodfellow, Ian and Pouget-Abadie, Jean and Mirza, Mehdi and Xu, Bing and Warde-Farley, David and Ozair, Sherjil and Courville, Aaron and Bengio, Yoshua},
  journal={Advances in neural information processing systems},
  volume={27},
  year={2014}
}

@inproceedings{gong2023redeem,
  title={Redeem myself: Purifying backdoors in deep learning models using self attention distillation},
  author={Gong, Xueluan and Chen, Yanjiao and Yang, Wang and Wang, Qian and Gu, Yuzhe and Huang, Huayang and Shen, Chao},
  booktitle={2023 IEEE Symposium on Security and Privacy (SP)},
  pages={755--772},
  year={2023},
  organization={IEEE}
}

@inproceedings{qu2019gan,
  title={GAN-DP: Generative adversarial net driven differentially privacy-preserving big data publishing},
  author={Qu, Youyang and Yu, Shui and Zhang, Jingwen and Binh, Huynh Thi Thanh and Gao, Longxiang and Zhou, Wanlei},
  booktitle={ICC 2019-2019 IEEE International Conference on Communications (ICC)},
  pages={1--6},
  year={2019},
  organization={IEEE}
}

@article{mirza2014conditional,
  title={Conditional generative adversarial nets},
  author={Mirza, Mehdi and Osindero, Simon},
  journal={arXiv preprint arXiv:1411.1784},
  year={2014}
}

@inproceedings{wang2019beyond,
  title={Beyond inferring class representatives: User-level privacy leakage from federated learning},
  author={Wang, Zhibo and Song, Mengkai and Zhang, Zhifei and Song, Yang and Wang, Qian and Qi, Hairong},
  booktitle={IEEE INFOCOM 2019-IEEE conference on computer communications},
  pages={2512--2520},
  year={2019},
  organization={IEEE}
}

@inproceedings{zhang2019poisoning,
  title={Poisoning attack in federated learning using generative adversarial nets},
  author={Zhang, Jiale and Chen, Junjun and Wu, Di and Chen, Bing and Yu, Shui},
  booktitle={2019 18th IEEE international conference on trust, security and privacy in computing and communications/13th IEEE international conference on big data science and engineering (TrustCom/BigDataSE)},
  pages={374--380},
  year={2019},
  organization={IEEE}
}

@article{fang2021privacy,
  title={Privacy preserving machine learning with homomorphic encryption and federated learning},
  author={Fang, Haokun and Qian, Quan},
  journal={Future Internet},
  volume={13},
  number={4},
  pages={94},
  year={2021},
  publisher={MDPI}
}

@article{wang2024goldfish,
  title={Goldfish: An Efficient Federated Unlearning Framework},
  author={Wang, Houzhe and Zhu, Xiaojie and Chen, Chi and Esteves-Ver{\'\i}ssimo, Paulo},
  journal={arXiv preprint arXiv:2404.03180},
  year={2024}
}

@inproceedings{samaria1994parameterisation,
  title={Parameterisation of a stochastic model for human face identification},
  author={Samaria, Ferdinando S and Harter, Andy C},
  booktitle={Proceedings of 1994 IEEE workshop on applications of computer vision},
  pages={138--142},
  year={1994},
  organization={IEEE}
}

@article{li2021neural,
  title={Neural attention distillation: Erasing backdoor triggers from deep neural networks},
  author={Li, Yige and Lyu, Xixiang and Koren, Nodens and Lyu, Lingjuan and Li, Bo and Ma, Xingjun},
  journal={arXiv preprint arXiv:2101.05930},
  year={2021}
}

@article{yao2024machine,
  title={Machine unlearning of pre-trained large language models},
  author={Yao, Jin and Chien, Eli and Du, Minxin and Niu, Xinyao and Wang, Tianhao and Cheng, Zezhou and Yue, Xiang},
  journal={arXiv preprint arXiv:2402.15159},
  year={2024}
}

@article{chen2023unlearn,
  title={Unlearn what you want to forget: Efficient unlearning for llms},
  author={Chen, Jiaao and Yang, Diyi},
  journal={arXiv preprint arXiv:2310.20150},
  year={2023}
}

@article{kurmanji2024towards,
  title={Towards unbounded machine unlearning},
  author={Kurmanji, Meghdad and Triantafillou, Peter and Hayes, Jamie and Triantafillou, Eleni},
  journal={Advances in neural information processing systems},
  volume={36},
  year={2024}
}

@article{jia2023model,
  title={Model sparsification can simplify machine unlearning},
  author={Jia, Jinghan and Liu, Jiancheng and Ram, Parikshit and Yao, Yuguang and Liu, Gaowen and Liu, Yang and Sharma, Pranay and Liu, Sijia},
  journal={arXiv preprint arXiv:2304.04934},
  volume={1},
  number={2},
  pages={3},
  year={2023}
}

\section*{Appendix}

\section*{A \ Preliminaries}
\customlabel{A}
\subsection*{A.1 \ Machine Learning}
Machine learning algorithms developed allow computers to learn and improve automatically based on data. This process typically involves statistical analysis of large datasets to identify patterns and trends \cite{jordan2015machine}. The design of these algorithms is often inspired by the essence of human learning, enabling them to adjust their behavior through experience. In supervised learning, algorithms learn a function from known input-output pairs intending to make accurate predictions on new, unseen data. Unsupervised learning, on the other hand, does not rely on labeled data but seeks to discover hidden structures and patterns within the data. Additionally, there are other learning paradigms such as semi-supervised learning and reinforcement learning, each applicable to different application scenarios and data types.

Taking supervised learning as an example, it requires a model to be trained on a dataset $D=\left\{\left(x_i,y_i\right)\right\}_{i=1}^n$, where each instance consists of an input feature vector $x_i$ and its corresponding label $y_i$. Here, $x_i\in\ X$ represents a feature vector from the input feature space, and $y_i\in\ Y$ represents a label value from the output label space. The goal of machine learning is to train a model $M$ with parameters $w$ to automatically learn the mapping between the input feature space and the output label space $X\rightarrow\ Y$. The model is trained by optimizing the loss function $\mathcal{L}\left(D,w\right)=\sum_{(x_i,y_i)\in D}\ l\left((x_i,y_i),w\right)$, which measures the discrepancy between the model's predicted label ${\hat{y}}_i=M\left(x_i,w\right)$ and the true label $y_i$ 
 . The optimal parameters $w^\ast$ for the model are obtained by optimizing the objective function equation \ref{优化目标}:
 \begin{equation}
    w^\ast=argmin\ L\left(D,w\right)
    \label{优化目标}
\end{equation}



\subsection*{A.2 \ Generative Adversarial Networks}
Generative Adversarial Networks (GANs) are a class of deep learning models that generate data through an adversarial training process, proposed by Ian Goodfellow and others in 2014. The core idea of GANs is to simultaneously train two models: a Generator and a Discriminator, which compete against each other during the training process to improve the quality of the generated data.

\textit{Generator}. The goal of the Generator is to create realistic data samples that cannot be distinguished by the Discriminator. It receives a random noise vector as input, typically sampled from a probability distribution, such as a normal distribution. The Generator transforms this noise vector into an output of the same dimension as the real data through a series of layers, usually convolutional or fully connected layers, and this output is the data sample we aim to generate.

\textit{Discriminator}. The task of the Discriminator is to distinguish between input data from the real dataset and fake data generated by the Generator. It is also a neural network that takes real or generated data as input and outputs a probability value indicating the likelihood that the input data is real. The Discriminator’s goal is to maximize its ability to correctly classify real and generated data.

\textit{Adversarial Training Process}. In the training process of GANs, the Generator and Discriminator adversarially interact. The Generator attempts to produce increasingly realistic data to deceive the Discriminator, while the Discriminator strives to improve its capability to identify authenticity. This process can be seen as a minimax game, where the Generator tries to minimize one objective function, while the Discriminator tries to maximize another. These two objective functions are usually opposite, hence they compete against each other.

\textit{Training Objective}. The training objective of GANs can be formalized as equation \ref{eq:gan_loss}:
\begin{equation}
\label{eq:gan_loss}
    \begin{split}
    \underset{G}{min}\,\underset{D}{max}\ V\left(D,G\right)=
E_{x\sim p_{data}\left(x\right)}\left[log{D\left(x\right)}\right]\\
    +E_{x\sim p_z\left(z\right)}\left[log{(1-D(G(z)))}\right]    
    \end{split}
\end{equation}
where $p_{data}\left(x\right)$ is the distribution of real data, $p_z\left(z\right)$ is the distribution of the noise vector, $G\left(z\right)$ is the output of the generator, and 
$D\left(x\right)$ is the probability output by the discriminator.

\section*{B \ Algorithms}
\customlabel{B}

The proposed federated unlearning approach, detailed in Algorithm \ref{alg:Unlearned}. 
The visible evaluation algorithm for assessing machine unlearning capability, detailed in Algorithm \ref{alg:training-phase} and \ref{Decision-making phase}.

\begin{algorithm}[!h]
    \caption{Federated Unlearning Approach}
    \label{alg:Unlearned}
    \SetAlgoLined
    \SetKwInOut{Input}{Input}
    \SetKwInOut{Output}{Output}

    \Input{local training dataset $D_i$, initialized model parameter $\omega_0$, learning rate $\mu$, deletion request $D_f^i$}
    \Output{unlearned global model}
    \SetKwFunction{Unlearn}{Unlearn}
    \SetKwFunction{LocalTraining}{LocalTraining}
    \SetKwProg{Server}{Server}{:}{}
    \SetKwProg{Users}{Users}{:}{}
    \SetKwProg{Procedure}{Procedure}{:}{}
    \Procedure{Efficient Federated Unlearning Approach}{}{
        Reinitialize global model and distribute to all users $i$: $\omega_i^0 \leftarrow \omega_0$\;
        \For{$t = 0, 1, \ldots, N$}{
            \Users{}{
            \ForEach{user $i$ in parallel}{
                \If{no deletion request}{
                    \LocalTraining($\omega_i^t$, $D_i$)\;
                }
                \Else{
                    \ForEach{unlearned user $i$ in parallel}{
                        $D_f^i\leftarrow$ deleted data\;
                        $\omega_t \leftarrow$\Unlearn($\omega_0$, $\omega_t$, $D_f^i$ )\;
                    }
                }
            }
            }
            
            \Server{}{
                Update global model parameters $\omega_{t+1}$\;
            }
        }
        \Return $\omega_{t+1}$\;
    }


    \Procedure{\Unlearn{$\omega_0$, $\omega_t$, $D_f^i$}}{
        Initialize teacher model $M_T$ parameters: $\omega_T \leftarrow \omega_0$\;
        Initialize student model $M_S$ parameters: $\omega_S \leftarrow \omega_t$\;
        \ForEach{local epoch $e=0,1,2,\ldots,n$}{
            \ForEach{batch $b\in D_f^i$}{
                    $\omega_T(e+1) \leftarrow \omega_T(e) - \mu\nabla l(M_T, M_S, b)$\;
            }
        }
        \Return $\omega_T(e+1)$\;
    }


    \Procedure{\LocalTraining($\omega_i^t$, $D_i$)}{
        \ForEach{local epoch $i=0,1,2,\ldots,n$}{
            \ForEach{batch $b\in D_i$}{
            $\omega_i^t(e+1) \leftarrow \omega_i^t(e) - \mu\nabla l(\omega_i^t, b)$\;
        }
        }
        
        \Return $\omega_i^t(e+1)$\;
    }

\end{algorithm}

\begin{algorithm}[]
    \caption{Training phase}
    \label{alg:training-phase}
    \SetAlgoLined
    \SetKwInOut{Input}{Input}
    \SetKwInOut{Output}{Output}
    \SetKwFunction{TrainDiscriminator}{TrainDiscriminator}
    \SetKwFunction{TrainGenerator}{TrainGenerator}
    \SetKwProg{TrainDiscriminator}{Train Discriminator}{:}{end}
    \SetKwProg{TrainGenerator}{Train Generator}{:}{end}
    \SetKw{KwFor}{for}
    \SetKw{KwTo}{to}
    \SetKw{KwEndFor}{end for}
    \SetKw{KwForBatch}{for batch}
    \SetKw{KwEndBatch}{end batch}
    \SetKw{KwForSteps}{for steps}
    \SetKw{KwEndSteps}{end}
    \SetKw{KwReturn}{return}
    \SetKwProg{Procedure}{Procedure}{:}{}
    \SetKwFunction{CalculateGeneratorLoss}{Generator Loss}
    \SetKwFunction{CalculateDiscriminatorLoss}{Discriminator Loss}

    \Input{unlearned classification model $C_u$, $\mu_d$, $\mu_c$, $\mu_f$, $\mu_r$, learning rates $\alpha_G$, $\alpha_D$, real dataset $D$}
    \Output{Generator $G$}
    \emph{Randomly initialize generator model} $G(\cdot, \omega_G)$, \emph{and discriminator model} $D(\cdot, \omega_D)$\;

    \For{$t = 0, 1, \ldots, N$}{
        \ForEach{$b \in D$}{
            \TrainDiscriminator{}{
                Obtain random noise $z$ and random label $y_r$\;
                Generate fake sample $\widetilde{b} \leftarrow G(z, y_r, \omega_G)$\;
                $l_D \leftarrow$ \CalculateDiscriminatorLoss($\mu_f, \mu_r, \widetilde{b}, b, G, D$)\;
                Update $\omega_D \leftarrow \omega_D - \alpha_D \nabla l_D$\;
            }
            \ForEach{$j \ steps$}{
                \TrainGenerator{}{
                    Obtain random noise $z$ and random label $y_r$\;
                    Generate fake sample $\widetilde{b} \leftarrow G(z, y_r, \omega_G)$\;
                    $l_G \leftarrow$ \CalculateGeneratorLoss{$\mu_d, \mu_c, \widetilde{b}, G, D, C_u$} \;
                    Update $\omega_G \leftarrow \omega_G - \alpha_G \nabla l_G$\;
                }
            }
        }
        
    }
    \Procedure{\CalculateGeneratorLoss{$\mu_d, \mu_c, \widetilde{b},  D, C_u$}}{
    Calculate the confidences of $\widetilde{b}$ using $D$ \;
    Calculate the predicted labels of $\widetilde{b}$ using $C_u$ \;
    
    Compute $l_G$ using equation \ref{生成器的目标函数}\;
    \Return $l_G$\;
    }
    \Procedure{\CalculateDiscriminatorLoss{$\mu_f, \mu_r, \widetilde{b}, b, D$}}{
    Calculate the confidences of $\widetilde{b}$ and $b$ using $D$ \;
    Compute $l_D$ using equation \ref{判别器损失}\;
    \Return $l_D$\;
    }
    
    
\end{algorithm}
\begin{algorithm}[]
\caption{Decision-making phase}
\label{Decision-making phase}

\DontPrintSemicolon 

\KwIn{Trained Generator weight $\omega_G$}
\KwOut{Decision}
Obtain random noise $z$ and all categories labels $y_{\text{all}}$;\\
Generate samples for all classes $b_{\text{all}} = G(z, y_{\text{all}}, \omega_G)$;\\
Assess the effectiveness of unlearning based on $b_{\text{all}}$;\\
\eIf{the quality of images corresponding to the unlearned category is poor}{
  $\text{Decision} = 1$;\\
}{
  $\text{Decision} = 0$;\\
}

\Return $\text{Decision}$;
\end{algorithm}

\begin{table*}[!t]
\centering 
\tabcolsep=0.386cm
\caption{Ablation Study of the Importance of Different Components of the Loss Function.}
\label{Ablation Study of the Importance of Different Components of the Loss Function}
\renewcommand\arraystretch{1.1}
\begin{tabular}{cccccc}
\hline
Epoch               & Metrics  & Total loss & w/o Distillation loss & w/o Attention map Alignment loss & w/o Hard loss \\ \hline
\multirow{2}{*}{5}  & backdoor$\downarrow$ & 1.58\%     & 0.63\%                & 7.91\%                           & 58.23\%       \\
                    & acc$\uparrow$      & 84.20\%    & 50.59\%               & 58.04\%                          & 88.70\%       \\
\multirow{2}{*}{10} & backdoor$\downarrow$ & 0.95\%     & 0.00\%                & 0.00\%                           & 74.05\%       \\
                    & acc$\uparrow$      & 77.37\%    & 30.52\%               & 28.57\%                          & 87.71\%       \\
\multirow{2}{*}{15} & backdoor$\downarrow$ & 0.00\%     & 0.00\%                & 0.00\%                           & 76.90\%       \\
                    & acc$\uparrow$      & 74.94\%    & 10.00\%               & 10.00\%                          & 86.32\%       \\
\multirow{2}{*}{20} & backdoor$\downarrow$ & 0.00\%     & 0.00\%                & 0.00\%                           & 84.49\%       \\
                    & acc$\uparrow$      & 66.28\%    & 10.00\%               & 10.00\%                          & 86.87\%       \\ \hline
\end{tabular}
\end{table*}

\section*{C \ Model Architectures}
\customlabel{C}
The discriminator model designed for the MNIST dataset is detailed as follows. 
It accepts single-channel 28$\times$28 pixel images as input.
The initial step involves passing the image through the first convolutional layer, comprising 32 filters of size 5$\times$5. To preserve the image dimensions, the output is padded with 2 and subsequently processed by the LeakyReLU activation function\cite{hitaj2017deep}.
After that, an average pooling layer reduces the feature map size by half to 14$\times$14. 
Following this, the second convolutional layer extracts further features using 64 filters of size 5$\times$5, employing the LeakyReLU activation function. 
Another average pooling layer halves the feature map size again to 7$\times$7. 
The resulting 7$\times$7 feature map is then flattened into a one-dimensional vector and passed through a fully connected layer.
The first fully connected layer reduces the feature dimension from 3136 to 1024, employing the LeakyReLU activation function. Following this, a second fully connected layer produces a scalar output, which is then transformed into a probability between 0 and 1 using a Sigmoid function. This probability indicates the likelihood that the input image is real data.

The corresponding generator model for the MNIST dataset is detailed below. 
It takes two inputs: 10-class conditional labels and 100-dimensional Gaussian noise.
The conditional label is first embedded into a 50-dimensional vector, followed by a linear transformation to reshape it into a 49-dimensional vector.
This vector is subsequently transformed using the ReLU activation function and reshaped to a 7$\times$7 feature map.
The Gaussian noise is initially processed through a linear layer to create a 6272-dimensional vector, which is then passed through the LeakyReLU activation function. 
This output is reshaped to a 128-channel 7$\times$7 feature map. 
Following the initial process, the core architecture consists of several transposed convolution layers. 
It starts by upsampling the feature map size from 7$\times$7 to 14$\times$14 using 128 filters, followed by another upsampling to 28$\times$28 with an additional set of 128 filters. 
The final output, a 28$\times$28 pixel handwritten digit image, is generated through a single-channel 3$\times$3 convolution layer followed by a Sigmoid activation function.

The discriminator architecture for the AT\&T dataset is designed as follows.
It starts with a 64$\times$64 pixel single-channel image.
It sequentially passes through three convolutional layers with 32, 64, and 128 filters of size 5x5, employing padding to preserve dimensions and LeakyReLU as an activation function.
Each convolutional layer is followed by average pooling that reduces the spatial size by half.
The output from the last pooling layer is flattened and passed through fully connected layers, {decreasing the dimensionality from 8192 to 1024}. Finally, a scalar probability indicating the authenticity of the image is generated by a Sigmoid function.

The generator for the AT\&T dataset is structured as follows. 
It takes 40-class conditional labels and 100-dimensional Gaussian noise as input. 
The conditional labels are first upsampled to a 1$\times$4$\times$4 feature map, and the noise is transformed into a 99$\times$4$\times$4 feature map. These two feature maps are then combined to form a 100$\times$4$\times$4 input feature map.
After that, it starts by upsampling to 8$\times$8 using 512 filters, followed by further upsampling to 16$\times$16, 32$\times$32, and finally 64$\times$64 with 256, 128, and 64 filters respectively. 
Each convolutional layer is accompanied by batch normalization and LeakyReLU activation. 
The final 128x128 feature map is generated by a transposed convolution with a single filter, followed by average pooling and hyperbolic tangent activation to produce facial images.

\section*{D \ Ablation Study of Unlearning Loss Function}
\customlabel{D}
To investigate the significance of different components of the loss function, we conducted an ablation study on the loss function. 
We trained a ResNet32 model on the CIFAR-10 dataset using four different combinations of loss functions: (1) total loss, which includes distillation loss, attention map alignment loss, and hard loss; (2) without distillation loss, which includes attention map alignment loss and hard loss; (3) without attention map alignment loss, which includes distillation loss and hard loss; and (4) without hard loss, which includes distillation loss and attention map alignment loss.
In the total loss configuration, each of the distillation loss, attention map alignment loss, and hard loss is weighted equally at 0.33. Conversely, in configurations where one of these losses is excluded, the weight for the remaining losses is set to 0.5. Detailed experimental results are presented in Table 1.
The experimental results are summarized in Table \ref{Ablation Study of the Importance of Different Components of the Loss Function}.

From the table, it can be observed that when the distillation loss is absent, although the backdoor attack success rate is low, the model accuracy drops significantly. Similarly, when the attention map alignment loss is lacking, there is also a sharp decline in model accuracy.
However, when the hard loss is absent, the model accuracy and backdoor success rate remain high.
After analyzing the results, we can infer that both the distillation loss and the attention map alignment loss significantly prevent the hard loss of the model. Moreover, in the absence of hard loss, the backdoor pattern can still be maintained within the model, failing to achieve the unlearning goal.



\end{document}